\documentclass{article}

\usepackage{PRIMEarxiv}

\usepackage[utf8]{inputenc} % allow utf-8 input
\usepackage[T1]{fontenc}    % use 8-bit T1 fonts
\usepackage{hyperref}       % hyperlinks
\usepackage{url}            % simple URL typesetting
\usepackage{booktabs}       % professional-quality tables
\usepackage{amsfonts}       % blackboard math symbols
\usepackage{nicefrac}       % compact symbols for 1/2, etc.
\usepackage{microtype}      % microtypography
\usepackage{lipsum}
\usepackage{fancyhdr}       % header
\usepackage{graphicx}       % graphics
\graphicspath{{media/}}     % organize your images and other figures under media/ folder

\usepackage{hyperref}
\usepackage{graphicx}
\usepackage{amsmath}
\usepackage{amssymb}
\usepackage{booktabs}
\usepackage{multirow}
\usepackage{placeins}
\newcommand{\quotes}[1]{``#1''}
\usepackage{graphics}
\usepackage{svg}

\usepackage[scleft,nofillcomment,ruled,vlined]{algorithm2e}
\usepackage{algorithmicx}
\usepackage[capitalize]{cleveref}
\crefname{section}{Sec.}{Secs.}
\Crefname{section}{Section}{Sections}
\newcommand{\etal}{\textit{et al.}}
\newcommand{\ie}{\textit{i.e.}}
\newcommand{\eg}{\textit{e}.\textit{g}.}
\usepackage[]{geometry}

\usepackage[normalem]{ulem}
\title{Leveraging Synthetic Data to Learn Video Stabilization Under Adverse Conditions
\thanks{The paper is under review.} 
}

\author{
  Abdulrahman~Kerim \\
  School of Computing and Communications\\
  Lancaster University, UK\\
  \texttt{a.kerim@lancaster.ac.uk} \\
  \And
  Washington L. S. Ramos \\
  Computer Science Department\\
Universidade Federal de Minas Gerais, Brazil\\
  \texttt{washington.ramos@dcc.ufmg.br} \\
   \AND
   Leandro Soriano Marcolino \\
     School of Computing and Communications\\
  Lancaster University, UK\\
   \texttt{l.marcolino@lancaster.ac.uk} \\
   \And
   Erickson R. Nascimento \\
     Computer Science Department\\
   Universidade Federal de Minas Gerais, Brazil\\
   \texttt{erickson@dcc.ufmg.br} \\
   \And
   Richard Jiang \\
     School of Computing and Communications\\
  Lancaster University, UK\\
   \texttt{r.jiang2@lancaster.ac.uk} \\
}

\begin{document}
\maketitle

\begin{abstract}
% The widespread availability of cameras and the growing popularity of video-sharing websites have led to the rapid growth of unedited and hard-to-watch videos.
Video stabilization plays a central role to improve videos quality. However, despite the substantial progress made by these methods, they were, mainly, tested under standard weather and lighting conditions, and may perform poorly under adverse conditions.
% they are limited to standard weather and lighting conditions, making them perform poorly under adverse conditions.
% In this paper, we propose to use synthetic data to provide robustness across different weather conditions when training a synthetic-aware algorithm for video stabilization.
In this paper, we propose a synthetic-aware adverse weather robust algorithm for video stabilization that does not require real data and can be trained only on synthetic data. We also present Silver, a novel rendering engine to generate the required training data with an automatic ground-truth extraction procedure. Our approach uses our specially generated synthetic data for training an affine transformation matrix estimator avoiding the feature extraction issues faced by current methods.
%In this paper, we propose to use a synthetic-aware algorithm for video stabilization, which is more robust than the state-of-the-art methods across different weather conditions. We use synthetic data for training an affine transformation matrix estimator, avoiding the feature extraction issues faced by current methods.
% In this paper, we propose a new algorithm for video stabilization, outperforming state-of-the-art results under adverse conditions.
% Our key idea is to use synthetic data for training an affine transformation matrix estimator, avoiding the feature extraction issues faced by current methods.
% In this paper, we propose to use only synthetic data to train a novel adverse weather condition-aware video stabilization method to achieve state-of-the-art results.
Additionally, since no video stabilization datasets under adverse conditions are available, we propose the novel VSAC105Real dataset for evaluation. We compare our method to five state-of-the-art video stabilization algorithms using two benchmarks. Our results show that current approaches perform poorly in at least one weather condition, and that, even training in a small dataset with synthetic data only, we achieve the best performance in terms of stability average score, distortion score, success rate, and average cropping ratio when considering all weather conditions. Hence, our video stabilization model generalizes well on real-world videos and does not require large-scale synthetic training data to converge.

% Our results show that we achieve the best performance in terms of stability average score, distortion score, success rate, and average cropping ratio.
% Our results show that our method achieved the best performance in terms of stability average score, distortion score, success rate, and average cropping ratio.

\end{abstract}

% keywords can be removed
\keywords{Video Stabilization\and Synthetic Data\and Affine Transformation}

\section{Introduction}
%Context Paragraph

%Problem why other methods do not work

%Our solution

%How are we beating other methods? + Contribution

%Context Paragraph

% === Establish the territory: video stabilization of unedited videos ===

Over the past several years, we have witnessed an explosion of videos being recorded and shared on the Internet. However, most shared videos are unedited and shaky, which makes them unpleasant to watch. Therefore, video stabilization techniques became an essential step in the video processing pipeline, gaining momentum as more unedited videos are being created and shared.
%Video stabilization is an essential technique in the field of computer vision. It has started to obtain more momentum as more videos are being recorded and shared on the Internet. Unpleasant shaky videos are not only hard to watch but can also cause computer vision models to fail. State-of-the-art video stabilization approaches perform well under standard conditions but struggle at adverse conditions. Collecting training videos in these conditions is hard, dangerous, and time-consuming. At the same time, creating photo-realistic synthetic videos simulating these conditions is rather hard, expensive, and poses certain limitations because of the domain gap problem between synthetic and real domains. Hence, we propose a novel synthetic-aware video stabilization algorithm that leverages synthetic data and achieves state-of-the-art results using only a small-scale synthetic dataset. Figure~\ref{fig:HoNet} depicts the main steps of this approach.
% Therefore, many efforts have been made to solve this task. %
% === Establish the niche: video stabilization recorded under adverse conditions ===
State-of-the-art video stabilization approaches perform well under standard conditions but struggle at adverse conditions. Furthermore, collecting training videos in these adverse conditions is hard, dangerous, and time-consuming.

%Problem why other methods do not work
%In general, most video stabilization approaches follow three steps: motion estimation, motion correction, and video synthesis \cite{guilluy2020video}. 
The training data bottleneck mentioned above causes many video stabilization methods to be essentially non-learning-based commonly adopting affine or homography matrix estimation in the camera motion estimation step, to extract the camera trajectory. Usually, features extraction, description, and matching are involved in this process. Feature extractors like SIFT~\cite{lowe2004distinctive} and the learning-based ones, like R2D2~\cite{r2d2} and ASLFeat~\cite{luo2020aslfeat}, can perform well under standard weather conditions and enough illumination, but they may fail under challenging conditions such as foggy, rainy, and snowy weather, as well as night-time scenes and abrupt changes of illumination. For instance, rain and snow drop particles and textureless scenes under fog or at night-time pose a clear challenge to find sufficient robust features. Failing to accurately estimate the actual camera trajectory, in the motion estimation step, causes the error to propagate to the later steps of the process, decreasing the quality of the stabilized video.
% Moreover, non-learning-based video stabilizers include many variables that must be tuned manually to work well in practice. These parameters include cropping window size, number of iterations to pass over the video, and sensitivity of the feature extractor, to name a few.

\begin{figure}[t]
    \centering
    \includegraphics[width=12cm]{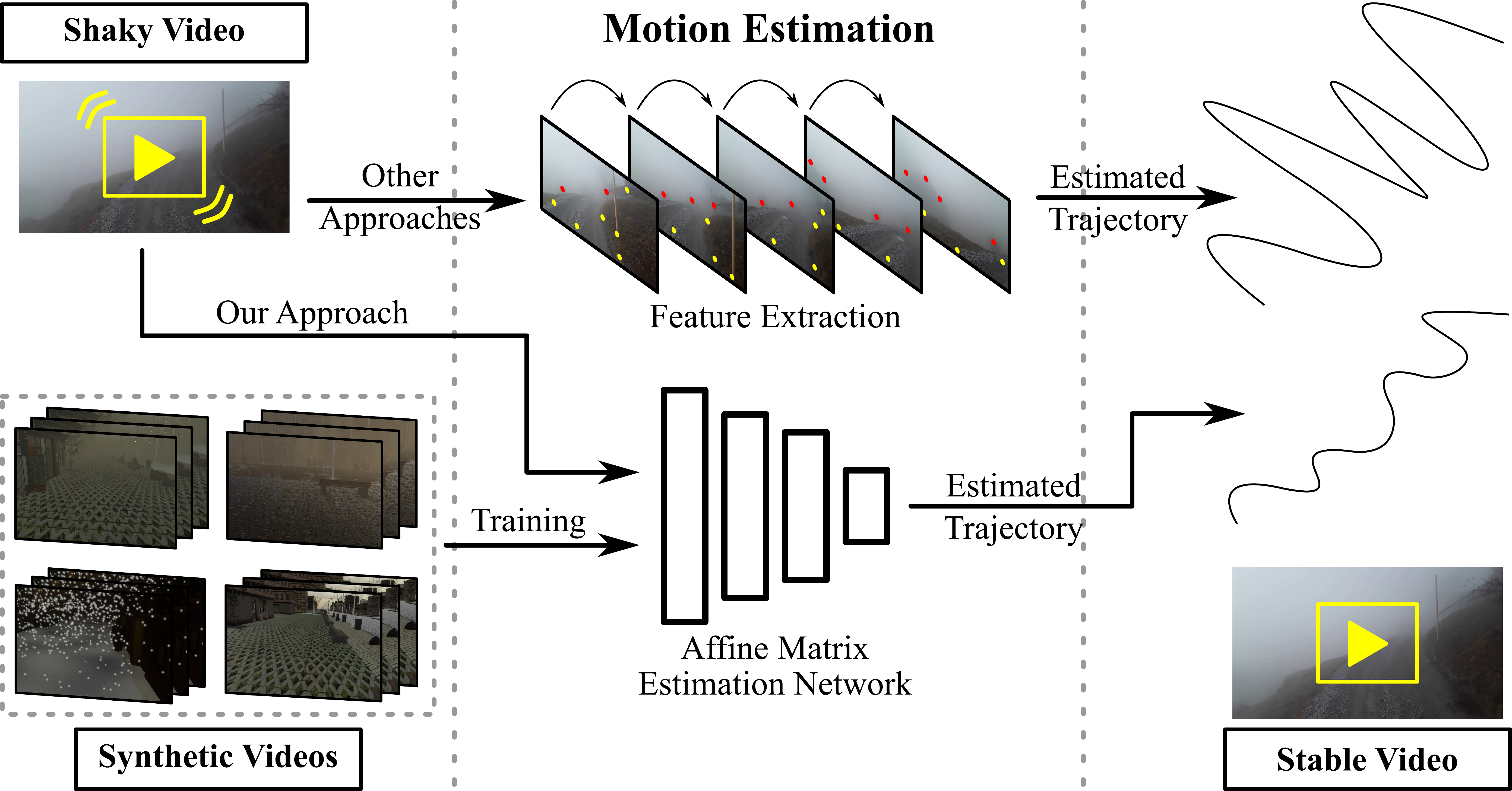}
    \caption{Our key idea is to use specially-designed synthetic data to train an affine transformation matrix estimation CNN.}
    \label{fig:HoNet}
\end{figure}

% === Occupy the niche: your wonderful approach :-) ===
Synthetic data have shown a great progress in the field of computer vision~\cite{liu2021urbanscene3d,kerim2021using,shafaei2016play,tsirikoglou2022synthetic}. Its increase in popularity started to attract many researchers to apply it for different computer vision problems. Most importantly, synthetic data seems a promising solution to overcome the lack of suitable data for training supervised learning models. However, we believe that the potential of synthetic data is not simply in the amount of available data for training.
 In fact, in this paper we propose a novel synthetic-aware video stabilization algorithm that leverages synthetic data and achieves state-of-the-art results using only a small-scale synthetic dataset. Figure~\ref{fig:HoNet} depicts the main steps of this approach. Leveraging the powerful tools of the Unity game engine, we built \textit{Silver}, which creates three-dimensional, photo-realistic virtual worlds procedurally at run-time. The system can automatically diversify many essential scene attributes like weather conditions, time of the day, crowdedness, to name a few. Most importantly, our system can generate the required ground-truth training data for our learning-based video stabilization method.
% Our system can generate ground-truth training data for our learning-based video stabilization method.
Our novel video stabilization model can be trained with a small number of synthetic videos. With no training or fine-tuning on real data, our model is more robust and accurate than the state-of-the-art across different weather conditions. To the best of our knowledge, this is the first work to study video stabilization under adverse weather conditions and utilize synthetic videos for video stabilization.
% With no training or fine-tuning on real data, our model can surpass the current state-of-the-art video stabilizers.

% The recent great success of Artificial Neural Networks (ANN) in solving complex problems motivated many researchers to apply it to machine perception problems too. 
% In parallel to this, the great advancement in chip design, microelectronics, and the introduction of General-Purpose Graphics Processor Architectures like the General Purpose Graphics Processing Unit (GPGPU), facilitated training deep neural networks with millions of parameters. 
% These deep ANNs with their high degree of non-linearity, help in approximating the real functions or phenomena behind complex vision problems, like video stabilization, in a much more accurate way. Unfortunately, training these deep learning models requires a great amount of data together with their corresponding annotations or ground-truths. Finding, collecting, and annotating suitable data is cumbersome, time-consuming, error-prone, expensive, and subject to privacy issues. Perhaps, the lack of diverse, high quality, and precisely labeled data for video stabilization task, can be attributed to the previously mentioned reasons. Additionally, the nature of the task makes it rather hard to acquire training data from real world. As two cameras need to capture the world from the same point of view while one is stable and the other is shaky. 

Despite supervised learning-based approaches~\cite{wang2018deep,liu2021hybrid,yu2020learning,ali2020learning} being able to learn parameters like cropping window, sensitivity, and even to extract discriminative features, there is no sufficient labeled data for obtaining high-quality results in any condition.
Finding, collecting, and annotating relevant data is cumbersome, time-consuming, error-prone, expensive, and subject to privacy issues. It is widely accepted that collecting many videos under adverse conditions is not easy since various attributes are needed to be diversified, such as camera motion, scene type, time of the day, scene crowdedness, and recording resolution. Even assuming that videos under these attributes can be captured, there is still a problem capturing training data for this task. Using two cameras, one with a mechanical stabilizer and another without, is still not an ideal solution since the difference between the two camera locations will cause the scene to be captured from two different viewpoints. Thus, learning-based models may not be able to learn well the video stabilization task, especially under adverse conditions.
% Thus, learning-based models will not be able to learn well the video stabilization task.

%Our solution
% To solve the aforementioned drawbacks, in this paper we propose a novel approach that removes the charge over feature extractors. 
%In this paper, we propose a novel approach that removes the charge over feature extractors. 
% A promising solution seems to be in the %hidden in the exceptional
% leading-edge game engines like Unity \cite{UnityGameEngine}, Unreal Engine \cite{UnrealGameEngine}, and CryEngine \cite{CryGameEngine}. 
%Leveraging the powerful tools of the Unity game engine, we built a system that creates three-dimensional, photo-realistic virtual worlds, procedurally at run-time. The system can diversify automatically many essential scene attributes like weather condition, time of the day, crowdedness and so on. Our system can generate ground-truth training data for our learning-based video stabilization method. Our novel video stabilization model can be trained with a small number of synthetic videos. With no training or fine-tuning on real data our model can surpass the current state-of-the-art video stabilizers.
% ~\cite{kerim2021silver}
Hence, our main contributions are three-fold: \noindent \textit{i)} a novel synthetic-aware video stabilization method achieving state-of-the-art results on real videos and being trained only on synthetic videos. It is worth noting that our key contribution is the idea of using specially designed synthetic
data with a simple yet powerful architecture for the task of video stabilization;
% for video stabilization able to generate an unlimited number of training videos for video stabilization under a wide set of attributes; %% Should we mention ground-truth?
\noindent \textit{ii)} a new synthetic training data generator, \textit{Silver}~\footnote{A preliminary version of \textit{Silver}~\cite{kerim2021silver} was presented as an unarchived workshop paper.}, which is able to generate an unlimited number of training videos for training our video stabilization algorithm under a wide set of attributes;
\noindent \textit{iii)} a new video stabilization dataset, the VSAC105Real, composed of real videos spanning foggy, rainy, and snowy weather conditions, and night-time attributes. %. We collect a novel dataset called VSAC105Real to assess the performance of video stabilization algorithms under adverse conditions.
% It is worth noting that our key contribution is not in terms of the architecture itself, however, in terms of providing a strong baseline and the idea of using specially designed synthetic data with a simple yet powerful architecture for the task of video stabilization.  
The implementation of our proposed video stabilization algorithm, our simulator, and the datasets are all available at \url{https://github.com/A-Kerim/SyntheticData4VideoStabilization}.
% \EN{Consider mentioning that the dataset, the code of your stabilization algorithm and simulator will be available to the community.}

\section{Related Work}
Our work utilizes synthetic data for the task of video stabilization. Thus, video stabilization and synthetic data generation literature is briefly reviewed in this section. 

\subsection{Video Stabilization}

Video stabilization methods can be categorized broadly into non-learning-based and learning-based approaches. 
Non-learning-based video stabilization methods do not perform training. For instance, Grundmann~\etal~\cite{grundmann2011auto} stabilize the shaky camera trajectory using L1-norm optimization under constraints.
% The work of Grundmann et al. \cite{grundmann2011auto} is an example of non-learning based video stabilization. They stabilize the shaky camera trajectory using L1-norm optimization technique under certain constrains.
% \sout{Their method minimizes the first, second, and third derivatives of the resulting camera path to compute a smooth camera trajectory composed of constant, linear, and parabolic segments}.
Similarly, 
% \EN{Drawbacks and benefits of using non-learning-base methods?} \EN{Only one reference in this category?}
% Thus, giving a more pleasant stabilized video.
Bradley~\etal~\cite{bradley2021cinematic} address the stabilization task as a constrained convex optimization problem. 
% \sout{They solve the homography estimation by working in log-homography space, where they model it without loss of convexity.}
These non-learning-based methods stabilize videos without the need for training data to tune the model's parameters. However, they tend to give less pleasant results, can work only under predefined conditions, and the model's parameters must be tuned manually.

The learning-based approaches, on their turn, can be classified into unsupervised and supervised approaches.
Non-supervised video stabilization methods require training videos but do not demand for annotated pairs of shaky-stable videos. 
Deep Iterative FRame INTerpolation (DIFRINT)~\cite{choi2020deep} is an unsupervised learning-based approach that can be trained in an end-to-end manner. It can stabilize videos without cropping original frames. Essentially, this method utilizes the frame interpolation technique to synthesize middle frames for video stabilization.  
% choi2020deep,wang2018deep
Supervised learning approaches, on the other hand, require human labeled ground-truth data, which is the main limitation of applying them to video stabilization.
%is the problem of generating ground truth data. 
Some works approached this problem using a mechanical stabilizer to generate ground-truth stable videos with their shaky counterparts. As an example, StabNet~\cite{wang2018deep} followed this procedure, generating videos for training CNNs for video stabilization. The network learns a warping transformation of multi-grids given the shaky video frames and the previously stabilized ones. On the other hand, Liu~\etal~\cite{liu2021hybrid} apply a learning-based hybrid-space fusion to compensate for optical flow inaccuracy. They synthesize stabilized frames by fusing the warped content estimated from neighboring frames. 
% \sout{They can stabilize frames without cropping them and with fewer visual artifacts.}
Yu~\etal~\cite{yu2020learning} learn to stabilize videos using optical flow. They compute the per-pixel warp field from the optical flow of the shaky video, allowing to better handle moving objects and occlusion.
% proposed using optical flow for video stabilization. They compute the per-pixel warp field from the optical flow of the shaky video. This method learns to stabilize videos using the optical flow allowing to better handle moving objects and occlusion. 
In contrast to the work of Yu~\etal~\cite{yu2020learning}, Ali~\etal~\cite{ali2020learning} propose a full-frame supervised video stabilization method that does not require optical flow. Their novel pipeline for dataset generation uses a linearly moving window on high-resolution images.

Despite the substantial progress made by these methods, they present a partial solution to the video stabilization problem since it is assumed that they will work under normal weather conditions and sufficient illumination. However, finding resilient features in adverse conditions is rather challenging. For example, rain particles, foggy weather conditions, and low illumination pose clear challenges to finding robust features. Thus, it leads to inaccurate motion estimation and low-quality video stabilization. At the same time, collecting diverse real-world videos for training video stabilization methods is cumbersome and time-consuming. Moreover, obtaining data with perfect ground-truth is not feasible given the physical limitation of recording the same scene from the same perspective via steady and shaky cameras.  

Our video stabilization method belongs to the supervised learning-based category. However, unlike other methods, we use only synthetic data for training. To the best of our knowledge, this is the first work to utilize synthetic videos for video stabilization. No pre-training or fine-tuning on real data is required, and by using only a small-scale training dataset, our method is more robust than state-of-the-art methods.

\subsection{Affine and Homography Transformation}
Estimating the affine or homography transformation between two images are two common approaches to align one image to another. For that aim, there are different ways to find these matrices  like applying a feature extractor (\eg, SIFT~\cite{lowe2004distinctive}, ORB~\cite{rublee2011orb}, SURF~\cite{bay2006surf}, and OAN~\cite{zhang2019learning}) and an outlier rejection algorithm (\eg, RANSAC~\cite{fischler1981random} and MAGSAC~\cite{barath2019magsac}). At the same time, there are unsupervised \cite{zhang2020content} and supervised \cite{detone2016deep} methods that estimate, specifically, the homography matrix.  

Although the traditional approach, \ie, feature extraction, does well at standard conditions, it performs poorly under challenging conditions such as adverse weather conditions and low illumination. Moreover, supervised approaches cannot reflect scene parallax~\cite{ye2021motion} and generating a suitable training data is rather hard. Unsupervised approaches tried to solve these problems but they still fail under large baseline alignment which makes it unpractical for applications such as image stitching and video stabilization under sharp camera movements. 

Unlike previous methods, our model learns the affine transformation in a supervised manner using specially generated synthetic training data for this task and it does not need any real data in training time. 
% \sout{Additionally, it can be trained on unlimited number of training samples covering wide set of challenging scenarios such as heavy snow, dense foggy, and low illumination.}
Estimating the affine transformation is of a special importance in video stabilization because it allows us to recover the camera translation, rotation, and scale.
% \sout{ Then, to apply other algorithms to smooth the camera transitions over the frames.} 
Although it is possible to decompose the homography matrix to extract these information, it is not accurate. Moreover, training a model to estimate the affine transformation is much easier compared to estimating the homography transformation. As have been discussed by other similar works \cite{grundmann2011auto,li2015dual}, although homography can better model the camera motion between frames for a limited number of frames, it starts to cause many artifacts like skew and perspective as the number of frames becomes larger. At the same time, higher degrees of freedom transformations (e.g., homography) overfit easily even with some regularization. Thus, utilizing homography transformation is both harder to train (more parameters and easier to overfit) and more subject to artifacts.

\subsection{Synthetic Data Generation} 
%synthetic data
% \sout{Although the exponential increase in the amount of digital data today makes data collection easier than before, manual labelling of large volumes of examples with high quality and accurate labels still requires too much effort and comes with a high cost. The use of synthetic data in the computer vision community has shown to be a promising solution to overcome the lack of suitable data for training supervised learning models. Its increase in popularity started to attract many researchers to propose novel algorithms to generate such datasets with their corresponding ground truths.} %Although special care needs to be taken to weigh each method’s advantages and disadvantages, they seem a promising solution to overcome the lack of suitable data for training supervised learning models.
The use of synthetic data in the computer vision community has shown to be a promising solution to overcome the lack of suitable data for training supervised learning models~\cite{butler2012naturalistic,richter2016playing,shafaei2016play,liu2021urbanscene3d,kerim2021using,tsirikoglou2022synthetic}.
Adapting a specific video game to generate synthetic data with its corresponding ground-truth for the task of semantic segmentation was presented by Richter~\etal~\cite{richter2016playing}, who modified the game Grand Theft Auto V for that purpose. 
% \sout{They have shown that integrating synthetic and organic datasets in the training process for semantic segmentation improves the overall performance considerably.} 
At the same time, another work by Shafaei~\etal~\cite{shafaei2016play} investigated the use of photo-realistic video games to generate synthetic data and their corresponding ground-truths for image segmentation and depth estimation.
% Conducting many experiments, they proved that training a Convolutional Neural Network (CNN) using only synthetic data achieves a similar performance as with training on real data. 
Using open-source animation movies was another approach discussed by Butler~\etal~\cite{butler2012naturalistic}, who showed how to obtain an optical flow large-scale dataset, MPI-Sintel, following a systematic and easy process. Dosovitskiy~\etal~\cite{dosovitskiy2017carla} presented CARLA, a popular open-source simulator widely used for autonomous driving research. 
% \sout{CARLA is built as an additional layer on Unreal Engine \cite{UnrealGameEngine}. It supports different weather types and two illumination conditions. The CARLA simulator can also provide ground-truth data for semantic segmentation and depth estimation tasks.}
Recently, the UrbanScene3D simulator was proposed by Liu~\etal~\cite{liu2021urbanscene3d} and developed for computer vision and robotics research. The system supports autonomous driving and flying research in different environments. 
% \sout{It is built on the Unreal Engine \cite{UnrealGameEngine} and AirSim \cite{shah2018airsim}}.
GANcraft is another work by Hao~\etal~\cite{hao2021gancraft} for generating photorealistic images. 

% \sout{GANcraft can produce consistent images from different viewpoints without the need for paired ground-truth real images. GANcraft also allows controlling certain features of the generated images by leveraging style-conditioning images.}

%%% === Erickson ===
%I think the following paragraph is very relevant and should be keep it.
%%% === Erickson ===
Despite the great advancements of the previous simulators, they present a partial solution for the data generation issue because of the lack of control on the environmental elements. They also fail to randomize the scene elements which will lead to some clear repetitions (to scene elements) when a large-scale dataset is required to be generated by these methods.
%In addition, integrating new elements or behaviours to the scene, like new 3D model, material, or texture, is hard or unattainable. Moreover, these systems are limited to specific computer vision tasks. Although these approaches are based on high quality and rich 3D virtual worlds, the failure of such methods to randomize the scene elements will lead to some clear repetitions (to scene elements) when a large-scale dataset is required to be generated by these methods. Procedural synthetic data generation offers an alternative to the previous solution.
In our simulator, we utilize procedural content generation to generate 3D virtual worlds. Additionally, our simulator can generate a special training data for the task of video stabilization.
%
%Unfortunately, 
% Despite the great advancements of the previous simulators, they present a partial solution for the data generation issue because of the lack of control on the environmental elements. In addition, integrating new elements or behaviours to the scene, like new 3D model, material, or texture, is hard or unattainable. Moreover, these systems are limited to specific computer vision tasks. Although these approaches are based on high quality and rich 3D virtual worlds, the failure of such methods to randomize the scene elements will lead to some clear repetitions (to scene elements) when a large-scale dataset is required to be generated by these methods. Procedural synthetic data generation offers an alternative to the previous solution.
% In our system, we utilize Procedural Content Generation (PCG) to generate 3D virtual worlds. Thus, it can generate a special training data for the task of video stabilization. 
In our experiments, we show that generating appropriate training data and creating a pipeline that utilizes synthetic data, can achieve superior results.
% \kr{Our model can still provide better results in real videos even though our engine does not generate state-of-the-art photo-realistic videos. Alternatively, it ensures adequate photo-realism sufficient to alleviate the domain shift. Additionally, our synthetic aware algorithm and our specially designed synthetic data both contribute to teaching the model to accurately estimate the affine transformation while not overfitting to the synthetic data distribution being trained on. Thus, it mitigates the domain gap and achieves satisfactory results on real data.  }
In fact, our algorithm can still provide better results in real videos even though our engine does not generate state-of-the-art photo-realistic videos. Additionally, our synthetic aware algorithm and our specially designed synthetic data both contribute to teaching the model to accurately estimate the affine transformation while not overfitting to the synthetic data distribution being trained on. Thus, it mitigates the domain gap and achieves satisfactory results on real data.
\section{Methodology}
% General description of the figure 2
% Then each block one subsection
% add equations
% loss function

\begin{figure*}[t]
    \centering
    \includegraphics[width=\textwidth]{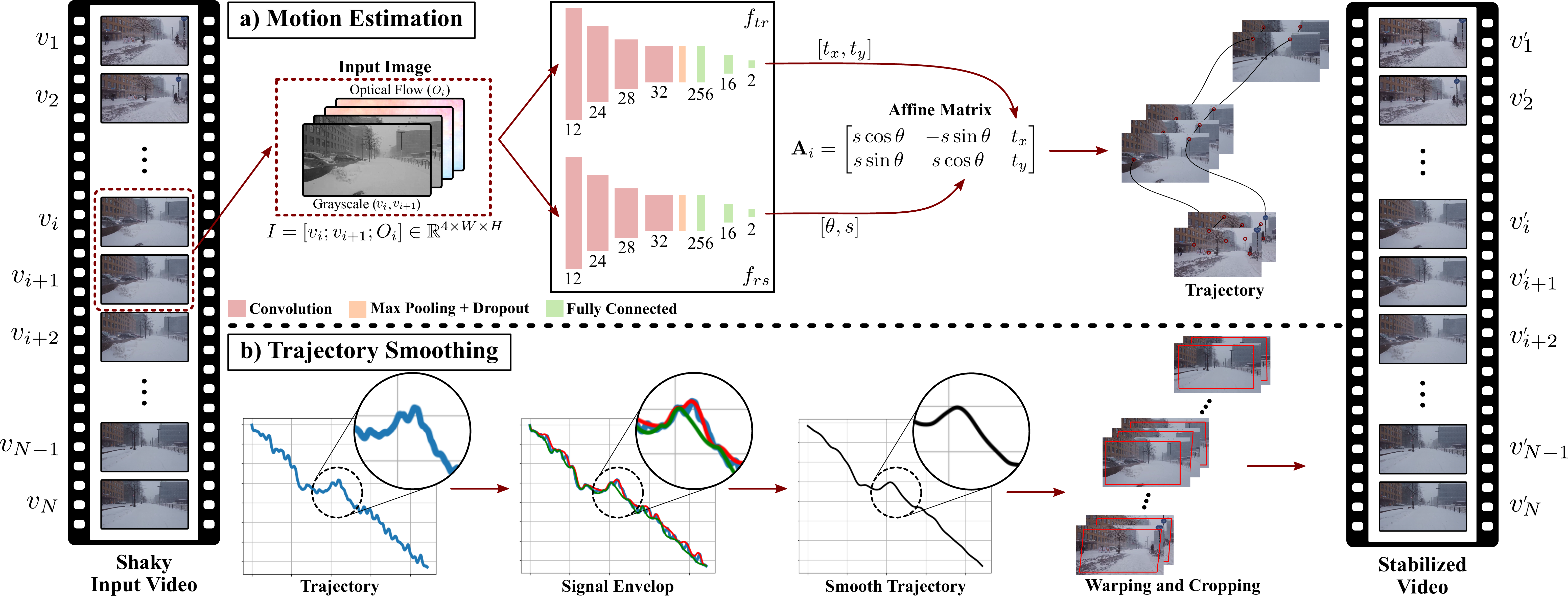}
    \caption{{\bf Video stabilization pipeline.} Our method estimates the translation, rotation, and scale for each pair of frames of the shaky video. After computing the camera trajectory, upper (red) and lower (green) bounds are found and averaged, and the Savitzky-Golay filter is applied to smooth the trajectory. Finally, warping and cropping is performed to generated the stabilized video. }
    \label{fig:pipeline}
\end{figure*}

Let ${V = \{v_1, v_2, \dots, v_N\}}$ be a shaky video composed of $N$ frames. The aim of our approach is to generate a stabilized version ${V^\prime = \{v^\prime_1, v^\prime_2, \dots, v^\prime_N\}}$ while preserving the original camera movement made by the recorder and maintaining its trend.
% \EN{while preserving the main intent of the recorder --- what do you mean with intent of the recorder? unclear}.
Our proposed method is composed of two major stages: \textit{i)} Motion Estimation; and \textit{ii)} Trajectory Smoothing. In the first stage, we train a motion estimation network using the ground-truth data from generated synthetic videos to estimate an affine transformation matrix $\mathbf{A}_i$ for every consecutive frames $v_i$ and $v_{i+1}$. Then, in the second stage, we calculate the camera trajectory ${\hat{T} = \{\hat{t}_1, \hat{t}_2, \dots, \hat{t}_N\}}$, where $\hat{t}_i = \sum_{j = 1}^{i} \hat{\mathbf{x}}_j$ and $\hat{\mathbf{x}}_j$ represents the estimated parameters for the pair of frames ${(v_j, v_{j+1})}$.  Following this and after smoothing $\hat{T}$, we warp and crop frames using the smoothed transformations retrieved from the smoothed trajectory  ${\tilde{T} = \{\tilde{t}_1, \tilde{t}_2, \dots, \tilde{t}_N\}}$. An outline of our approach is described in Figure~\ref{fig:pipeline}. %The first stage was trained in a supervised fashion using generated synthetic videos.  In a nutshell, our approach is described in Figure \ref{fig:pipeline}. 

% \subsection{Homography matrix estimation}
\subsection{Motion Estimation} 
% what is the input and output define the variables X
% Architecture howmany layers, functions, 
% training pipeline

The first stage of our pipeline consists in estimating the camera motion throughout the video. Most existing 2D-based stabilization approaches apply key-point feature extraction and tracking to solve this task~\cite{lee2009video,grundmann2011auto,liu2012video}. However, both feature extraction and tracking may fail under adverse weather conditions due to repetitive textures and partial occlusions caused by rain and snow particles or textureless objects under foggy weather and at night-time. To overcome this issue and properly recover the camera motion in $V$, we propose estimating parameters $t_x$, $t_y$, $\theta$, and $s$ of an affine transformation matrix
% $\mathbf{A}$
%%% === Erickson ===
% I removed the matrix definition to include the paragraph in RW about the drawbacks of the simulators
%%% === Erickson ===
% %%
\begin{equation}
    \mathbf{A} =
    \begin{bmatrix}
    s \cos{\theta} & -s\sin{\theta} & t_x \\
    s \sin{\theta} & s\cos{\theta} & t_y
\end{bmatrix} %i \in \{1,2,\dots, N\}
\end{equation}
for every consecutive pairs of frames using deep neural networks with synthetic data. Thus, we abdicate the feature extraction procedure entirely since, using our proposed engine, we can generate the ground-truth affine transformation needed for training as described in Section~\ref{sec:ground_truth_generation}.

Two identical networks are utilized: the ${f_{tr} : \mathbb{R}^{4 \times W \times H} \rightarrow \mathbb{R}^{2}}$ network that estimates the $x$ and $y$ translations, \ie, ${\mathbf{x}^{tr} = [t_x, t_y]}$; and ${f_{rs} : \mathbb{R}^{4 \times W \times H} \rightarrow \mathbb{R}^{2}}$  that outputs the rotation angle and scale values ${\mathbf{x}^{rs} = [\theta, s]}$, where $W=H=256$ is the center cropped image width and height. It is important to note that $f_{tr}$ and $f_{rs}$ share the same architecture, but not the same weights.
%We describe the networks architecture and training procedure 

%We use two identical networks \EN{identical means same architecture and same weights?} for estimating the parameters, one for the $x$ and $y$ translations that outputs  ${\mathbf{x}^{tr} = [t_x, t_y]}$, and another for the rotation angle and scale that returns ${\mathbf{x}^{rs} = [\theta, s]}$. %We describe the networks architecture and training procedure next. 
Both networks consist of a feature extractor implemented as four convolutional layers followed by a pooling and a dropout layer, and a regressor, which is a fully connected network composed of three linear layers that process the extracted features to estimate the parameters. The number of output channels of each layer is illustrated in Figure~\ref{fig:pipeline}-a. 
% Each network consists of a feature extractor ${f_e : \mathbb{R}^{4 \times W \times H} \rightarrow \mathbb{R}^{e}}$ implemented as four convolutional layers followed by a pooling and a dropout layer, and a regressor ${f_r : \mathbb{R}^{e} \rightarrow \mathbb{R}^{2}}$ which is a fully connected network that processes the extracted features to estimate the parameters (see Figure~\ref{fig:pipeline}-a). 
For each training step, we feed the networks with an input ${I = [v_i; v_{i+1}; O_i]} \in \mathbb{R}^{4 \times W \times H}$, where ${v_i, v_{i+1} \in \mathbb{R}^{W \times H}}$ are two consecutive grayscale frames from the input video $V$ and ${O_i \in \mathbb{R}^{2 \times W \times H}}$ is the optical flow map for the pair $(v_i, v_{i+1})$ obtained via the FlowNet2~\cite{IMKDB17,flownet2-pytorch}. Then, we estimate the parameters for $\mathbf{A}_i$ as 
% \begin{equation}
%     {(\hat{t}_x, \hat{t}_y) = f_{tr}(I)},
% \end{equation}
% \begin{equation}
%     {(\hat{\theta}, \hat{s}) = f_{rs}(I)}.
% \end{equation}
${\hat{\mathbf{x}}^{tr} = f_{tr}(I)}$ and ${\hat{\mathbf{x}}^{rs} = f_{rs}(I)}$. %We present more details about the architecture in the supplementary materials.

To optimize the parameters of the networks $f_{tr}$ and $f_{rs}$, we train separately each one using the Mean Squared Error (MSE) loss functions:
\begin{equation}
    \mathcal{L}_{tr} = \frac{1}{M}\sum_{m=1}^{M}(\hat{\mathbf{x}}^{tr} - \mathbf{x}^{tr})^2~\mathrm{and}~
    \mathcal{L}_{rs} = \frac{1}{M}\sum_{m=1}^{M}(\hat{\mathbf{x}}^{rs} - \mathbf{x}^{rs})^2,
\end{equation}
\noindent where $M$ is the number of training samples in a batch of randomly selected consecutive pair of frames to compose $I$. $\mathbf{x}^{tr}$ and $\mathbf{x}^{rs}$ are the ground-truth parameters of the affine matrix $\mathbf{A}_i$.

A key contribution of our approach is the usage of  specially-designed synthetic data to learn affine transformation.  Let $\mathcal{P}_i = \{\mathbf{p}_1, \mathbf{p}_2, \dots, \mathbf{p}_K \}$ denote the 2D coordinates of $K$ mark points at the frame $v_i$ from a generated synthetic video. Since these mark points in frame $v_i$ and $v_{i+1}$ are static in the world space, we can compute an affine transformation $\mathbf{A}_i$ with $4$ degrees of freedom using $\mathcal{P}_i$ and  $\mathcal{P}_{i+1}$. Following this strategy, we get the ground-truth values $t_x$, $t_y$, $\theta$, and $s$ for a single pair of images. The detailed process of the ground-truth data generation is described in Section~\ref{sec:ground_truth_generation}.

Finally, with the estimated parameters $\hat{\mathbf{x}} = [\hat{t}_x, \hat{t}_y, \hat{\theta}, \hat{s}]$ for each image pair in the video, we can compute the estimated camera trajectory ${\hat{T} = \{\hat{t}_1, \hat{t}_2, \dots, \hat{t}_{N-1}\}}$, where $\hat{t}_i = \sum_{j = 1}^{i} \hat{\mathbf{x}}_j$,
%%
% \begin{equation}
%     \hat{t}_i = \sum_{j = 1}^{i} \hat{\mathbf{x}}_j
%     \label{eq:camera_trajectory}
% \end{equation}
%%
% \noindent
and $\hat{\mathbf{x}}_j$ represents the estimated parameters for the pair of frames ${(v_j, v_{j+1})}$. 
It is important to note that, similar to~Grundmann~\etal~\cite{grundmann2011auto}, we do not use $\hat{t}_i$ directly to warp the shaky images; we warp the frames by applying  smoothed affine transformation composed using the smoothed translation, ration, and scale parameters as detailed in the following section.

\subsection{Trajectory Smoothing}
After estimating the shaky camera trajectory, we need to smooth it. In contrast to other methods like the work of Grundmann~\etal~\cite{grundmann2011auto}, where they tackle the camera trajectory smoothing as an optimization problem, we deploy the Savitzky-Golay filter~\cite{savitzky1964smoothing} on the averaged envelop of the shaky camera trajectory to smooth it, as described in the sequel.
%First, the envelop of the camera trajectory is calculated by finding local extrema and applying quadratic interpolation. Then, Savitzky-Golay filter~\cite{savitzky1964smoothing} is applied to smooth abrupt changes in the trajectory. Savitzky-Golay filter is applied because it removes noisy shakiness introduced by unintended camera movements without distorting the signal tendency. Thus, preserving the main camera movement. This approach allows our video stabilization method to preserve the global sense of the camera movement while removing unwanted shakiness.

\paragraph*{Signal Envelop Calculation} 
Given the camera trajectory $\hat{T}$, we first calculate the extremes of $\hat{T}$ by applying the first-order discrete derivative. Then, we interpolate the trajectory maxima ($\hat{T}_{max}$) and minima ($\hat{T}_{min}$) values to extract the upper and lower envelop, respectively. We empirically experimented with linear, cubic, and quadratic interpolations. However, quadratic interpolation presented the best results in our experiments since it makes smooth interpolations and tends to stay within the ranges of the interpolation points.
% Quadratic interpolation presented the best results since it makes smooth interpolation and stays within the range of interpolation points.
The final upper and lower signal envelops are represented as ${E_{up} = \{e^{up}_1, e^{up}_2, \dots, e^{up}_{N-1}\}}$ and ${E_{low} = \{e^{low}_1, e^{low}_2, \dots, e^{low}_{N-1}\}}$, respectively.

\paragraph*{Smoothing} After obtaining the upper and lower signal envelops, we apply the Savit\-zky-Golay filter~\cite{savitzky1964smoothing} on the average envelop ${\bar{E} = (E_{up}+E_{low})/2}$ to remove the unwanted sudden camera shakiness and create the smooth camera trajectory $\tilde{T} = \{\tilde{t}_1, \tilde{t}_2, \dots, \tilde{t}_{N-1}\}$ as shown in \mbox{Figure~\ref{fig:pipeline}-b}. The Savitzky-Golay filter smooths the digital signal by fitting a low-degree polynomial to consecutive signal points using linear least squares. This strategy has an advantage over other techniques as it preserves the signal tendency. Thus, $\tilde{T}$ still maintains the properties of $\hat{T}$ while ensuring smooth camera transition over time. After that, we calculate the difference between both trajectories $\delta_{T} = \hat{T} - \tilde{T}$. Then, the smoothed affine transformation parameters  $\tilde{X}$ can be calculated as $\tilde{X} = \hat{X} - \delta_{T}$, where $\tilde{X} = \{\tilde{\mathbf{x}}_1, \tilde{\mathbf{x}}_2, \dots, \tilde{\mathbf{x}}_{N-1}\}$ and $\hat{X} = \{\hat{\mathbf{x}}_1, \hat{\mathbf{x}}_2, \dots, \hat{\mathbf{x}}_{N-1}\}$, with $\tilde{\mathbf{x}}_i$ being the smoothed parameters $\tilde{t}_x$, $\tilde{t}_y$, $\tilde{\theta}$, and $\tilde{s}$.

% smoothTransforms = transforms - difference
%Taking the average signal envelop Savitzky-Golay filer is applied to remove unwanted sudden camera shakiness. Savitzky-Golay filer smooths the digital signal by fitting low-degree polynomial to consecutive signal points using linear least squares. This approach has an advantage over other techniques as it preserve the signal tendency. Thus, the smoothed camera trajectory $\tilde{T}$ still maintains the property of  $T$ while ensuring smooth camera transition over time.  

\paragraph*{Warping and Cropping} At last, we warp the video frames and crop them to compose the final video. 
For each video frame $v_i$, we compute its warped version $\tilde{v}_i$ by applying a transformation matrix to every pixel.
Formally, we retrieve $\tilde{\mathbf{x}}_i$ from the smoothed transformations $\tilde{\mathbf{X}}$ and use the smoothed parameters $\tilde{t}_x$, $\tilde{t}_y$, $\tilde{\theta}$, and $\tilde{s}$ to compose the smoothed affine matrix~$\tilde{\mathbf{A}}_i$.
% Formally, we retrieve $\tilde{t}_i$ from the smoothed camera trajectory $\tilde{T}$ and use the smoothed parameters $\tilde{t}_x$, $\tilde{t}_y$, $\tilde{\theta}$, and $\tilde{s}$ to compose an affine matrix $\tilde{\mathbf{A}}_i$. Then, for each frame $v_i$, we compute its warped version $\tilde{v}_i$ by applying the transformation matrix to every pixel.
Finally, we crop the warped frames using a predefined virtual cropping window similar to Grundmann~\etal~\cite{grundmann2011auto} to generate the stabilized video.

% \kr{}
% \WR{Kerim, please describe the cropping process (percentage of cropping?, etc.)} to obtain the final stable frame $v^\prime_i$.
%Given $\tilde{T}$, for each two consecutive frames  $v_i$ and $v_{i+1}$ the smooth transform $\tilde{T}_i$ is retrieved where ${\tilde{T}_i = \{\tilde{t}_x^i,\tilde{t}_y^i,\tilde{\theta}_i,\tilde{s}_i\}}$. After that, using  $\tilde{Tx_i},\tilde{Ty_i},\tilde{\theta_i}$ and $\tilde{S_i}$ the new affine transformation matrix is calculated and used to warp \(v_{i+1}\). Finally, the frames are cropped and the stabilized video is generated.
% If the hypothetical object becomes out of camera view or exceeds its time duration, it is removed. Each object is given a unique identifier over its lifetime.
\begin{figure*}[t]
    \centering
    \includegraphics[width=\textwidth]{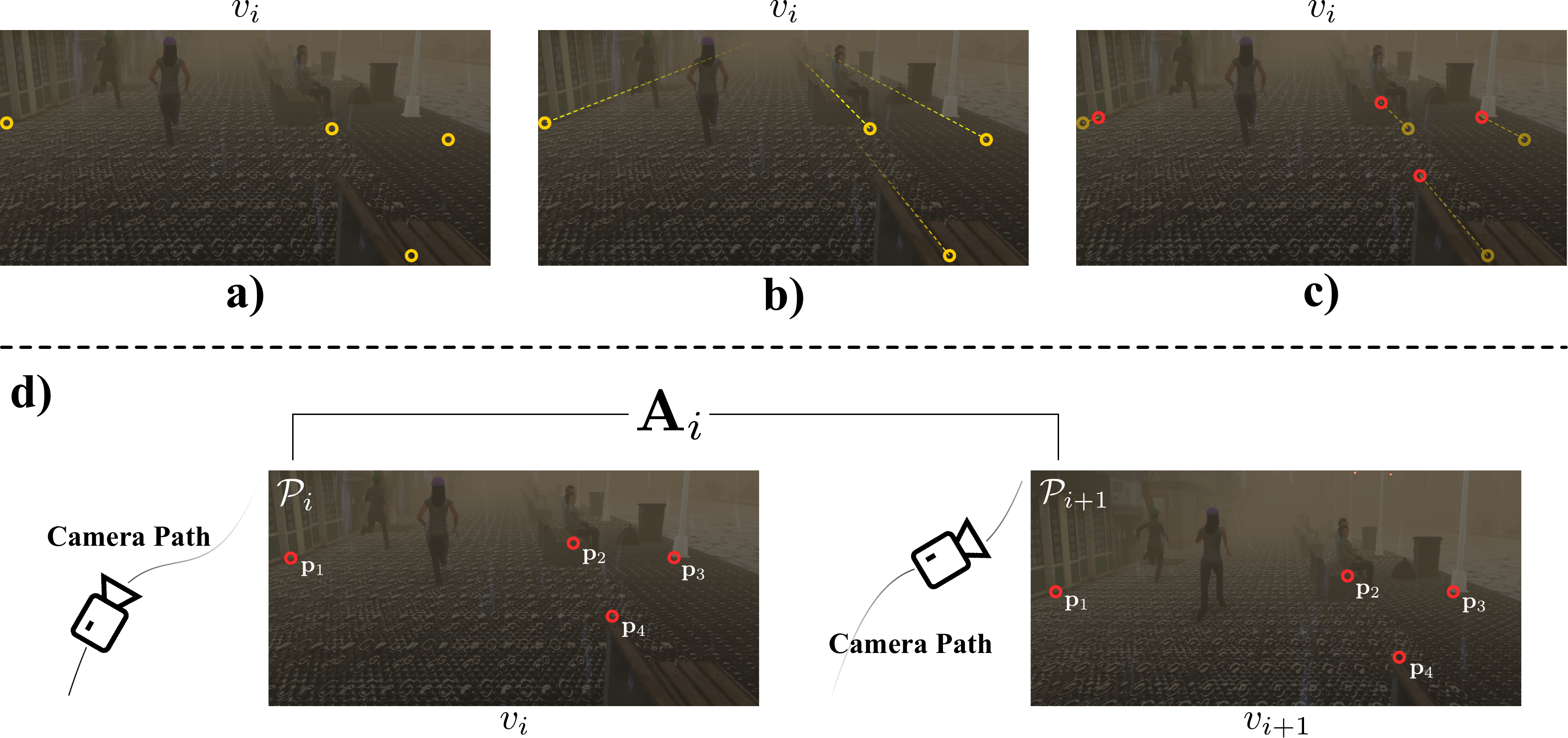}
    \caption{\textbf{Ground-truth Generation.} $K$ points are randomly sampled from the screen space (yellow circles in a). From each of these points we cast rays to infinity in the 3D scene space (dashed yellow lines in b), and create hypothetical objects at the intersection of these rays with the scene (red circles in c).We obtain the affine transformation matrix $A_i$ using the coordinates of the hypothetical objects in screen space $\mathcal{P}_i$ and $\mathcal{P}_{i+1}$ since they remain stationary in the scene from the frame $v_i$ to $v_{i+1}$.}

    \label{fig:groundtruth}
\end{figure*}
\subsection{Ground-truth Generation}
\label{sec:ground_truth_generation}
The goal of our network is to infer the affine transformation matrix given two consecutive frames. Thus, we need to generate the affine transformation ground-truth to supervise the training process. Our idea is two-fold: \textit{a)} create stationary hypothetical labeled objects in the 3D world scene; \textit{b)} record their coordinates in the screen space of the recording camera. In that way, we can guarantee that the coordinates of these objects in frames ${(v_i, v_{i+1})}$ correspond to exactly the same static elements in the 3D world seen by the recording camera at frames $v_{i}$ and $v_{i+1}$. In other words, we create a number of  invisible objects and save their coordinates in the camera space for each frame. In order to do so, for each frame a number of random points are sampled from the screen camera space. Then, a ray is cast from each of these points to infinity. At each ray's intersection point with the scene, a hypothetical invisible object is created. The object remains stationary for a number of seconds before being destroyed. For each frame, the object position in world space is transferred to the camera space and saved in XML format. Figure \ref{fig:groundtruth} demonstrates how these hypothetical objects are created.
% Extensible Markup Language (XML) format.

If the current camera view does not include the hypothetical object, or if it exceeds its time limit duration $\beta$, it is removed. Each object is given a Unique Identifier (UID) over its lifetime. Later in the post processing stage, for each two consecutive frames ${(v_i, v_{i+1})}$ using the UIDs of these hypthetical objects and their screen locations, the affine transformation ground-truth is calculated for each pair ${(v_i, v_{i+1})}$ as described in Algorithm \ref{algorithm:groundtruth}. It should be noted that generating millions of hypothetical objects is still a valid option. However, our algorithm provides a solution to create only objects in the field view of the recording agent improving the overall performance.

\begin{algorithm}[htb]
    \small
    \SetKwInOut{Input}{Require}
    \SetKwInOut{Output}{Ensure}
    \Input{$N_{frames}$, $T$}
    \Output{ Per frame $XML~file$ containing camera screen locations of a number of hypothetical objects stationary in scene space.   }
    $N_{frames};$ \Comment{Total Number of Frames to Generate}\\
    $T;$ \Comment{Sampling Period}\\

    % $Points$ $\leftarrow$ Sample $K$ points from camera screen space\;
    % $n_{hypothObjects}$ $\leftarrow$ Number of Active Hypothetical Objects\;
    \While{Recording}
    {
       \For{$ID_{frame} = 0;\ ID_{frame} < N_{frames};\ ID_{frame}{+}{+}$}
        {
            \uIf{$ID_{frame} \% T==0$}{
                $Points$ $\leftarrow$ $samplePoints($K$);$ \Comment{Sampling $K$ points on camera screen space}
                
                \ForEach{ point $\in$ $Points$ } {
                    % $n_{hypothObjects}$ =  $n_{hypothObjects} + 1$ $\leftarrow$ Increment counter \;
                     $M$ $\leftarrow$ $castRayToInfinity($point$);$ \Comment{Cast a ray from $point$ to infinity}\\
                    $point_{inters}$ $\leftarrow$ $findIntersPoint(M);$ \Comment{Intersection point between the ray with scene objects}\\
                    
                    $O$ $\leftarrow$ $createHyptheticalObject(point_{inters});$\\
                    $O.UID$ $\leftarrow$ $assignObjectUID(O);$ \Comment{Assign $O$ a unique identifier}\\
                    $O.ScreenPos \leftarrow Cam.WorldToScreen(O.WorldPos);$ \Comment{Transfer coordinates from world to screen space}
                    
                    \While{ $O$ is visible and did not exceed its lifetime }
                    { $Save(O.ScreenPos);$\\
                      $WaitFewFrames();$ }
                    $Destroy(O);$
                    
                }
            }
            \Else{$WaitFewFrames();$
            }
        }
    }
\caption{Affine Transfromation Ground-truth Generation}
\label{algorithm:groundtruth}
\end{algorithm}

\section{Silver: Framework for Generating Synthetic Data for Computer Vision Tasks}
\label{Silver}
There are many photo-realistic synthetic data generators like CARLA~\cite{dosovitskiy2017carla} and UrbanScene3D~\cite{liu2021urbanscene3d} that can be used to simulate photo-realistic, diverse, and visually complex 3D worlds. However, generating special data in such engines is cumbersome and most importantly, they do not support generating training data for video stabilization. \textit{Silver}, on the other hand, fills the gap and generates the required training data for this task. Additionally, it supports other computer vision tasks like semantic segmentation, instance segmentation, depth estimation, pose estimation, surface normals estimation. However, in this paper, we limit our discussion to the usability of our engine for video stabilization task.

% In this work, we show that more vital than photo-realistic and diverse 3D scenes is designing computer vision models targeted at using synthetic data.
In this work, we show that more vital than photo-realistic and diverse 3D scenes is designing computer vision models targeted at using synthetic data, and generating the appropriate synthetic data for these models. In virtual worlds, it is not only easy to control all scene aspects but also to generate more suitable training data for supervised learning algorithms.

%there is a factor that is more vital than photo-realism and diversity of the 3D scenes which is designing computer vision models targeted at using synthetic data. In virtual worlds it is not only easy to control all scene aspects but also to generate more suitable training data for supervised learning algorithms.
% \subsection{Synthetic Data Generation}

\begin{figure*}[t]
    \centering
    \includegraphics[width=\textwidth]{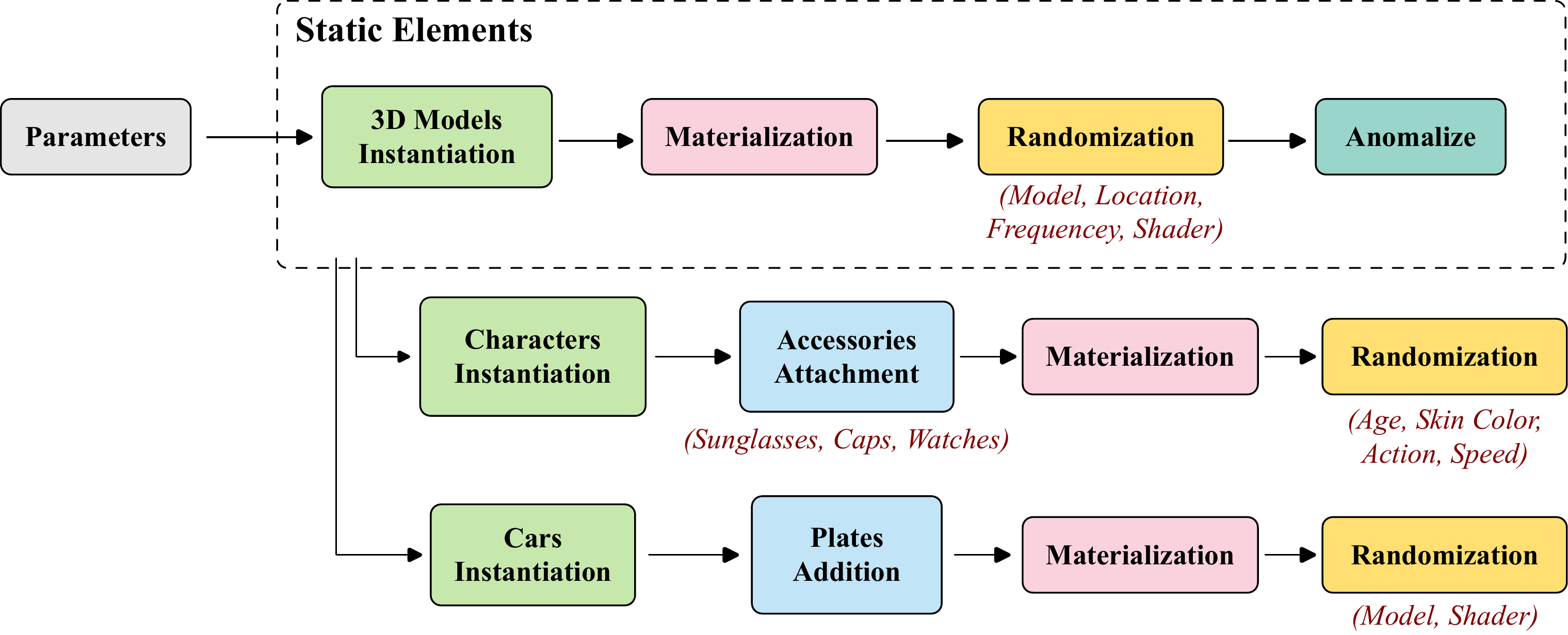}
    \caption{Flowchart describing the scene creation in \textit{Silver}.}
    \label{fig:flowchart}
\end{figure*}

\paragraph*{Simulator} Our work deploys synthetic data to train a synthetic-aware video stabilization algorithm. For that aim, we developed \textit{Silver} using the Unity game engine to generate our synthetic training datasets VSAC65Synth and VSNC35Synth. We employ the Procedural Content Generation (PCG) concept to create a full 3D virtual world at run-time while the system's extensibility is attained by taking advantage of the modular approach followed as we built the system from scratch. Although our simulator can provide clean, unbiased, and large-scale training and testing data for various computer vision tasks, we focus on the video stabilization task.
A shaky synthetic video is recorded after procedurally creating a 3D virtual world sampled from a predefined set of 3D models, materials, and animations. Note that for each video, a new virtual world is created to diversify the training data. A simplified flowchart describing the scene creation process is shown in Figure~\ref{fig:flowchart}.

\small{
\begin{figure*}[t]
    \centering
    \includegraphics[width=\textwidth]{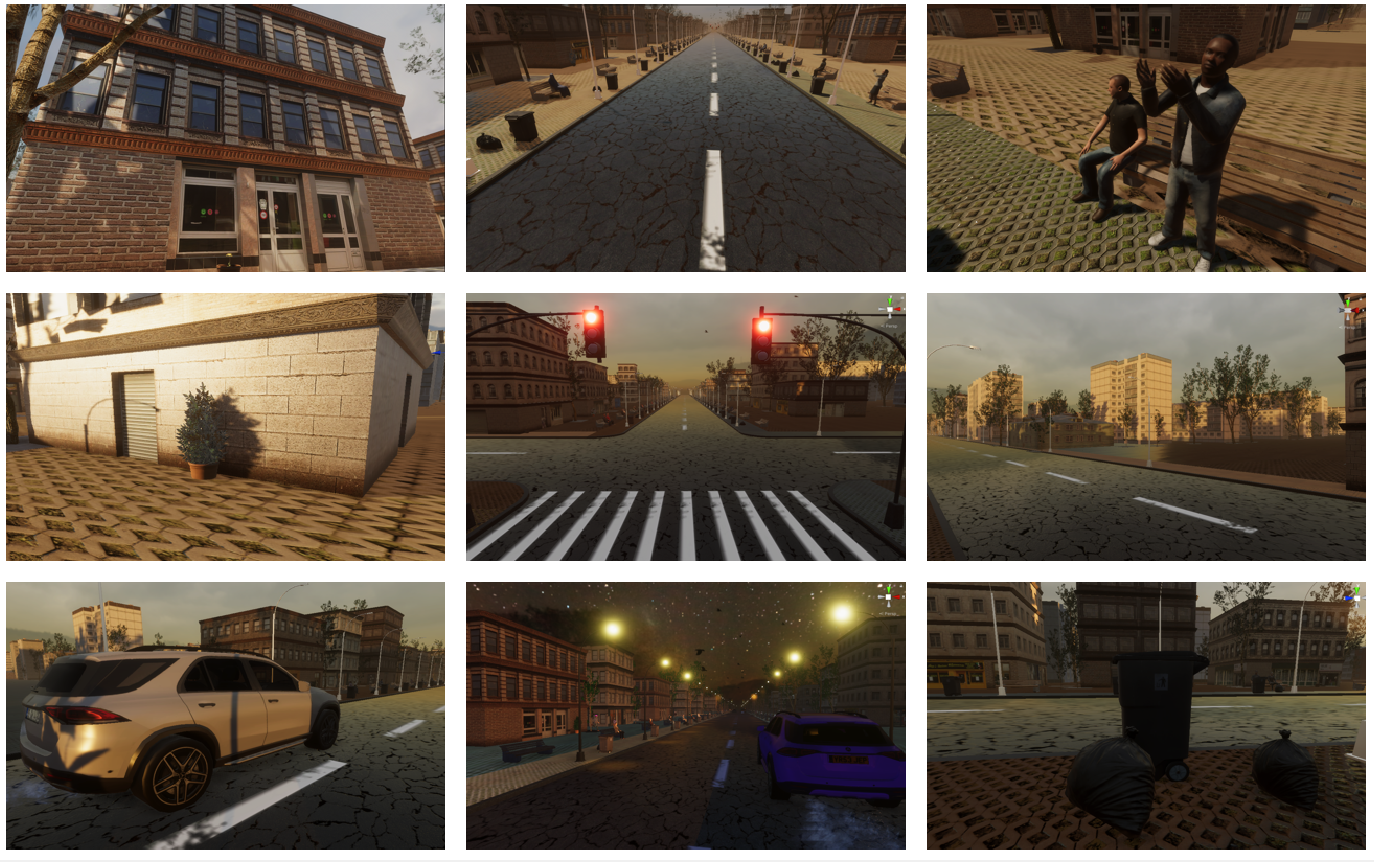}
    \caption{Samples from the procedurally generated scenes using \it{Silver}.}
    \label{fig:photo-realism}
\end{figure*}
}
\paragraph*{Static Elements} Starting from the given parameters, \textit{Silver} initially creates the static part of the 3D virtual world. In this part, first the street length and the number of crosses are set at random.
Following this, the buildings are created where buildings’ locations,
types, and frequency are set at random. After that, the other scene
elements like benches, trash containers and bags, trees, and other
elements are created. To further improve the realism and diversity of the generated scenes, we introduce a new variable called \textit{Anomaly Rate}; higher values will cause more artifacts to appear in the scene such as more street lights being off at night, more being on at daytime, and some trash bags being on roads. 

\paragraph*{Dynamic Elements} Once the static part of the scene is completed, the dynamic part of the scene is initiated. Initially, the characters generator retrieves the locations of buildings and benches, and instantiates characters based on the required characters density.
The Microsoft Rocketbox Avatar Library \cite{gonzalez2020rocketbox} is used to define the
character avatar, and the animations are selected based on character pose (standing or sitting). Character animations were adopted from Mixamo. In a similar way, the cars are created. However, number of cars and models are selected at random. Additionally, car shader attributes: Smoothness, Metallic, and BaseColour are all randomized at run-time to give different visual appearance even to the same car model. After that, the plates of the cars are selected at random from a large set collected manually from the web. The main processes are summarized in Figure~\ref{fig:flowchart}.
In parallel to that, the first-person videos are recorded using Cinemachine camera behaviour from Unity, attached to an AI navigation agent. We use Cinemachine since it gives unlimited sets of behaviours that enrich the diversity of the generated synthetic data in terms of the camera view angle and transition. 

\paragraph*{Camera Shakiness} To introduce shakiness to the recording camera, we create noise by using a predefined noise profile asset. The amplitude and frequency of the noise are randomly sampled from a uniform distribution. The noise
% To introduce shakiness to the recording camera, a simple algorithm 
% \EN{which algorithm? should be briefly described here}
is applied to change the translation and rotation of the recording agent camera. Figure~\ref{fig:photo-realism} demonstrates examples of the generated scenes.

\begin{figure*}[t]
    \centering
    \includegraphics[width=\textwidth]{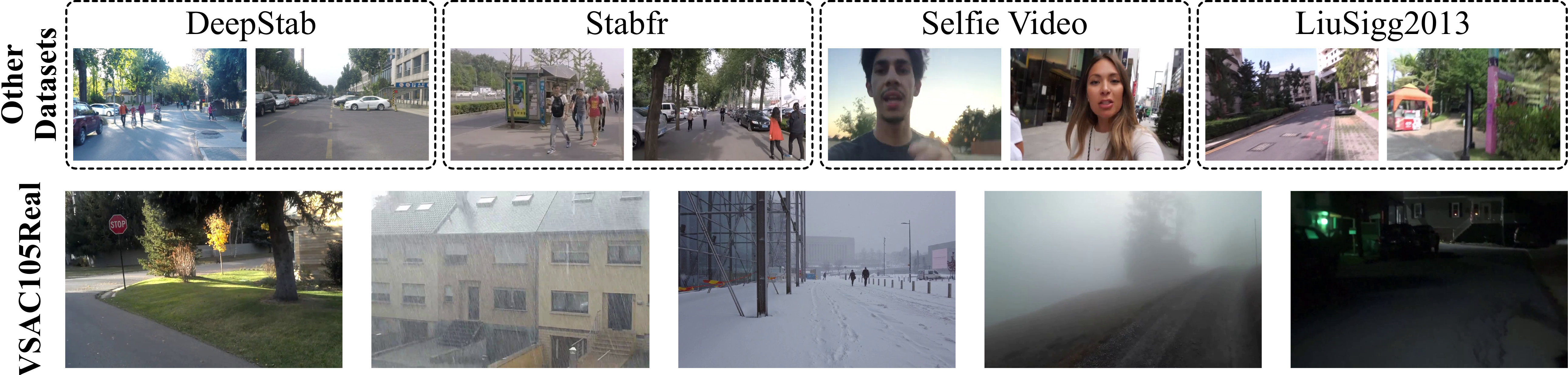}
    \caption{\textbf{VSAC105Real versus other datasets.} On top, each dashed box shows frames from other datasets. Our proposed dataset, VSAC105Real, is at the bottom. It includes more diverse and challenging attributes as compared to the other datasets.}
    \label{fig:VSAC105Real_diversity}
\end{figure*}

\small{
\begin{table}[t]
\centering
\caption{{\bf Dataset statistics.} Comparison among the available video stabilization datasets and VSAC105Real dataset.}
\label{table:statistics_datasets}
\begin{tabular}{lccc}
\toprule
Dataset Name & \#Videos & \begin{tabular}[c]{@{}c@{}}Average \\ \#Frames\end{tabular} & \begin{tabular}[c]{@{}c@{}}Total \\ \#Frames\end{tabular} \\ \midrule
DeepStab~\cite{wang2018deep}        & $61$  & $714$ & $43{,}585$ \\
Stabfr~\cite{zhang2018full}         & $45$  & $471$ & $21{,}200$ \\
Selfie Video~\cite{yu2018selfie}    & $33$  & $251$ & $8{,}308$  \\
LiuSigg2013~\cite{liu2013bundled}   & $144$ & $578$ & $83{,}257$ \\ \midrule
VSAC105Real                          & $105$  & $737$ & $77{,}477$ \\ \bottomrule
\end{tabular}

% \begin{tabular}{lccccccc}
% \toprule
% Dataset Name &
%   \#Videos &
%   \begin{tabular}[c]{@{}c@{}}Avgerage \\ \#Frames\end{tabular} &
%   \begin{tabular}[c]{@{}c@{}}Median \\ \#Frames\end{tabular} &
%   \begin{tabular}[c]{@{}c@{}}Min \\ \#Frames\end{tabular} &
%   \begin{tabular}[c]{@{}c@{}}Max\\  \#Frames\end{tabular} &
%   \begin{tabular}[c]{@{}c@{}}Sum\\ \#Frames\end{tabular} &
%   \begin{tabular}[c]{@{}c@{}}Std\\  \#Frames\end{tabular} \\ \midrule
% DeepStab        & 61  & 714 & 727 & 67  & 2,264 & 43,585  & 353 \\ 
% Stabfr          & 45  & 471 & 500 & 150 & 700  & 21,200  & 87  \\ 
% Selfie Video    & 33  & 251 & 261 & 134 & 397  & 8,308   & 59  \\ 
% LiuSigg2013 & 144 & 578 & 487 & 140 & 2,872 & 83,257  & 330 \\ \midrule
% % All             & 71  & 503 & 493 & 67  & 2872 & 156350 & --  \\ \hline \hline
% VSAC105Real           & 77  & 807 & 773 & 168  & 4,796 & 62,179 & 514  \\ \bottomrule
% \end{tabular}
\end{table}
}

\section{Real and Synthetic Datasets}
\subsection{Real Data Collection}
The available video stabilization benchmarks such as DeepStab~\cite{wang2018deep}, Stabfr~\cite{zhang2018full}, Selfie Video~\cite{yu2018selfie}, and LiuSigg2013~\cite{liu2013bundled} exclusively contain videos under normal weather condition and at a sufficient illumination. To assess the performance of the state-of-the-art video stabilization methods under foggy, rainy, snowy, and night-time conditions, we created the VSAC105Real dataset. 

Our dataset is composed of videos collected from YouTube using search queries like \quotes{Fog}, \quotes{Rain}, \quotes{Snow}, \quotes{Night}, \quotes{Adverse}, and \quotes{Severe}. We manually inspected all the videos and selected the ones with shaking camera movement. Then, we cut the videos to ensure continuous temporal criteria and the query attribute. VSAC105Real dataset comprises $105$ videos spanning normal, rainy, foggy, snowy, and night-time attributes. The first four attributes were selected to study the effect of severe weather conditions on video stabilization quality. Similarly, the night-time was chosen to understand the effect of low illumination on video stabilization. Table~\ref{table:statistics_datasets} shows a comparison among different video stabilization datasets and VSAC105Real dataset. The VSAC105Real dataset has the advantage in terms of the average number of frames. Moreover, it includes a diverse set of challenging attributes where videos are evenly distributed across the classes, \ie, $21$ videos per class. A visual comparison among VSAC105Real and other video stabilization datasets is depicted in Figure~\ref{fig:VSAC105Real_diversity}.

% Was removed from CVPR to ECCV because we have even distribution of classes which can be described easily in the text. Thus, no need for the table!!
% \begin{table}[t]
% \centering
% \caption{Classes distribution in VSAC105Real dataset.}
% \label{table:VSAC105Real_dist}
% \begin{tabular}{lccccc}
% \toprule
% Class & Fog & Night & Normal & Rain & Snow  \\
% \cmidrule{2-6}
% \#Videos & $21$ & $21$ & $21$ & $21$  & $21$ \\ \bottomrule
% \end{tabular}
% \end{table}

\subsection{Synthetic Data}
Using \textit{Silver}, we generate two different synthetic training datasets: VSNC35Synth under normal weather conditions and VSAC65Synth under both normal and adverse weather conditions. 
% The former, VSNC35Synth, is used for training our baseline and the later, VSAC65Synth, is used only in the ablation experiments.
% $A_{time} \sim \{day, night\}$,  $A_{Crowdedness} \sim \mathcal{N}(50,10)$...
%  $A_{weather} \sim \{sunny, cloudy, partly~cloudy\}$,
% $A_{weather} \sim \{sunny, cloudy, partly~cloudy, rainy, foggy, snowy\}$,

\paragraph{VSNC35Synth dataset} It is used in all training experiments, unless otherwise specified. It includes $35$ videos at $24$ fps and $400$ average number of frames per video; it covers only videos in normal weather conditions. The average number of frames was set to $400$ to match the available real video stabilization dataset average number of frames.  
\paragraph{VSAC65Synth dataset} It is used in one of the ablation study's experiments (Table~\ref{table:ablation}, under \textit{More Data} column). It consists of $65$ videos spanning normal, rainy, foggy, and snowy weather conditions at daytime and nigh-time. It has the same fps and average number of frames as the VSNC35Synth dataset.
\section{Experiments}

% \sout{In this section, we present the experimental setup we used to conduct our experiments and the comparison results.}
%  time limit duration
\subsection{Experimental Setup}
\paragraph*{Implementation Details}
We trained our method using \textbf{only} the synthetic data provided by our simulator, \ie, VSNC35Synth. We used the Adam optimizer~\cite{kingma2014adam} with a learning rate ${\alpha = 1e\mathrm{-}4}$ with ${\beta_1 = 0.9}$, ${\beta_2 = 0.999}$, and ${\epsilon = 1e\mathrm{-}8}$. Additionally, the hypothetical objects time limit duration was $\beta = 1$ second. We trained the ${t}_x$ and ${t}_y$ translation prediction model ($f_{tr}$) and rotation~${\theta}$ and scale ${s}$ prediction model ($f_{rs}$) for $65$ and $2$ epochs, respectively, using batches of size ${M = 40}$. After $10$ epochs, we decrease the learning rate of $f_{tr}$ to ${1e\mathrm{-}5}$. Our architecture is fully implemented in PyTorch, and our whole training procedure takes about $33$ hours on a Tesla V100 GPU. For the smoothing step, we used a window length, \ie, number of coefficients, equal to $51$ with $1$\textsuperscript{st} order polynomial as parameters to the Savitzky-Golay filter. We experimentally selected these values since they give slightly better results than others.
% $\hat{\mathbf{x}} = [\hat{t}_x, \hat{t}_y, \hat{\theta}, \hat{s}]$
%Our architecture was implemented in PyTorch. We used Adam optimizer~\cite{kingma2014adam} with a learning rate ${\alpha = 1e\mathrm{-}4}$ with ${\beta_1 = 0.9}$, ${\beta_2 = 0.999}$, and ${\epsilon = 1e\mathrm{-}8}$. After $10$ epochs, the translation network's learning rate, $\alpha_{trans}$, is decreased to ${\alpha_{trans} = 1e\mathrm{-}5}$. The translation network was trained for $65$ epochs while the rotation-scale network was trained for $2$ epochs. We used a batch size $40$. The training of our model took about $33$ hours on an NVIDIA GeForce GTX 1080 Ti graphics card. 

\paragraph*{Video Stabilization Methods}
We evaluate five state-of-the-art video stabilization algorithms on two datasets: \ie, VSAC105Real and Selfie Video~\cite{yu2018selfie}. The baselines span non-learning based~\cite{grundmann2011auto}, supervised~\cite{wang2018deep}, and unsupervised~\cite{choi2020deep} video stabilization methods. Furthermore, we compare our method to the one proposed by Yu~\etal~\cite{yu2020learning}, which heavily relies on optical flow since we also use optical flow. Similarly, we consider the work of Liu~\etal~\cite{liu2021hybrid} because they also use CNNs for video stabilization close to our work.

\setlength{\tabcolsep}{0.25em}
\small{
\begin{table*}[t!]
	\centering
	\caption{\textbf{Comparison across different weather conditions in the VSAC105Real dataset.} Our method presents the best average values in comparison to the other competitors for all metrics. \textbf{Bold} indicates the best and  \underline{underline} second best.}
% 	\EN{Consider including the citation each each method in the table.}
	\label{tab:results}
			
% 	\resizebox{0.99\linewidth}{!}{%

		\begin{tabular}{@{}clccccccc@{}}
			\toprule %\hline
			\multirow{3}{*}{\bf Metric} & \multirow{3}{*}{\bf Method} & & \multicolumn{5}{c}{{\bf Weather Condition}} & \multirow{3}{*}{\bf Average}\\
			\cmidrule{4-8} %\cmidrule{13-17} 
			& \multicolumn{1}{r}{} & & {\centering Fog} & {\centering Night} & {\centering Normal} & {\centering Rain} & {\centering Snow} &  \\ \midrule
			
			\multirow{5}{*}{\rotatebox[origin=c]{90}{\parbox[c]{2.5cm}{\centering Stability\\Avg. Score~$\uparrow$}}} 
			& FuSta~\cite{liu2021hybrid} & & $0.226$ & $0.683$ & $0.715$ & $0.679$ & $0.824$ & $0.626$ \\

			& Grundmann~\etal~\cite{grundmann2011auto} & & $0.642$ & $0.549$ & $0.620$ & $0.580$ & $0.809$ & $\underline{0.640}$ \\

			& StabNet~\cite{wang2018deep} & & $0.201$ & $0.469$ & $0.620$ & $0.577$ & $0.753$ & $0.524$ \\

			& DIFRINT~\cite{choi2020deep} & & $0.121$ & $0.212$ & $0.321$ & $0.247$ & $0.446$ & $0.270$ \\
			
			& Yu~\etal~\cite{yu2020learning} & & $0.401$ & $0.682$ & $0.665$ & $0.572$ & $0.834$ & $0.631$ \\

			& Ours & & $0.606$ & $0.619$ & $0.728$ & $0.687$ & $0.835$ & $\mathbf{0.695}$ \\

			\midrule
			
			\multirow{5}{*}{\rotatebox[origin=c]{90}{\parbox[c]{2.5cm}{\centering Distortion\\Score~$\ast$}}} 
			& FuSta~\cite{liu2021hybrid}           & & $0.202$ & $0.692$ & $0.725$ & $0.712$ & $0.798$ & $0.626$ \\

			& Grundmann~\etal~\cite{grundmann2011auto} & & $0.740$ & $0.617$ & $0.762$ & $0.667$ & $0.952$ & $\underline{0.748}$ \\

			& StabNet~\cite{wang2018deep}         & & $0.111$ & $0.518$ & $0.790$ & $0.597$ & $0.710$ & $0.545$ \\

			& DIFRINT~\cite{choi2020deep}         & & $0.367$ & $0.219$ & $0.351$ & $0.270$ & $0.476$ & $0.337$ \\
			
			& Yu~\etal~\cite{yu2020learning}        & & $0.372$ & $0.729$ & $0.654$ & $0.593$ & $0.804$ & $0.631$ \\

			& Ours            & & $0.719$ & $0.746$ & $0.952$ & $0.809$ & $0.997$ & $\mathbf{0.845}$ \\

			\midrule
			
			\multirow{5}{*}{\rotatebox[origin=c]{90}{\parbox[c]{2.5cm}{\centering Cropping\\Ratio~$\uparrow$}}}
			& FuSta~\cite{liu2021hybrid}           & & $0.286$ & $0.810$ & $0.905$ & $0.800$ & $0.950$ & $\underline{0.751}$ \\

			& Grundmann~\etal~\cite{grundmann2011auto} & & $0.759$ & $0.618$ & $0.760$ & $0.663$ & $0.948$ & $0.750$ \\

			& StabNet~\cite{wang2018deep}         & & $0.278$ & $0.579$ & $0.875$ & $0.667$ & $0.850$ & $0.650$ \\

			& DIFRINT~\cite{choi2020deep}         & & $0.399$ & $0.234$ & $0.392$ & $0.280$ & $0.490$ & $0.359$ \\
			
			& Yu~\etal~\cite{yu2020learning}        & & $0.476$ & $0.810$ & $0.842$ & $0.650$ & $0.947$ & $0.745$ \\

			& Ours            & & $0.762$ & $0.760$ & $0.945$ & $0.820$ & $0.999$ & $\mathbf{0.857}$ \\
			
			\midrule
			
			\multirow{5}{*}{\rotatebox[origin=c]{90}{\parbox[c]{2.5cm}{\centering Success\\Rate~$\uparrow$}}} 
			& FuSta~\cite{liu2021hybrid}           & & $0.280$ & $0.816$ & $0.905$ & $0.762$ & $0.905$ & $0.734$ \\

			& Grundmann~\etal~\cite{grundmann2011auto} & & $0.762$ & $0.619$ & $0.762$ & $0.667$ & $0.952$ & $\underline{0.752}$ \\

			& StabNet~\cite{wang2018deep}         & & $0.238$ & $0.524$ & $0.667$ & $0.571$ & $0.810$ & $0.562$ \\

			& DIFRINT~\cite{choi2020deep}         & & $0.429$ & $0.238$ & $0.381$ & $0.286$ & $0.476$ & $0.362$ \\
			
			& Yu~\etal~\cite{yu2020learning}        & & $0.480$ & $0.815$ & $0.762$ & $0.619$ & $0.857$ & $0.707$ \\

			& Ours            & & $0.765$ & $0.762$ & $0.950$ & $0.814$ & $1.000$ & $\mathbf{0.858}$ \\
			
			%\midrule
			& & & \multicolumn{6}{r}{\scriptsize{\textit{$\uparrow$Higher is better} \textit{$^\ast$Better closer to $1$}}} \\

			\bottomrule
		\end{tabular}
		
% 	}
% \vspace{-1mm}
\end{table*}
}
% Similarly, we consider the work of Liu et al. [9] because they also use CNNs for video stabilization
\paragraph*{Evaluation Metrics}
% Add equations [TODO]
To evaluate our approach, we use three metrics commonly used to evaluate video stabilization algorithms~\cite{liu2021hybrid,choi2020deep,wang2018deep,yu2020learning}: \noindent \textit{i) Stability Score}. It assesses the smoothness of the stabilized video; the higher the value the better. It is computed as the average between Stability Average Translation and Stability Average Rotation Scores. To compute this score, we estimate the homography matrix between $v_i$ and $v_{i+1}$ to obtain the translation and rotation arrays. Following this, we calculate their Fast Fourier Transform (FFT). Finally, we obtain the score by calculating the ratio between the $2$\textsuperscript{nd} through $6$\textsuperscript{th} frequency components and all frequency components. Note that the $0$\textsuperscript{th} frequency component is neglected;
\noindent \textit{ii) Distortion Score.} It measures the global distortion caused by a given video stabilization method. It fits a homography matrix between the original and stabilized videos. Then, it finds the anisotropic scaling among these frames; the closer to $1$, the better;
% it measures the global distortion by fitting a homography between the input and stabilized frames.
% it measures the anisotropic scaling of the homography between the input and output frames
\noindent \textit{iii) Cropping Ratio.} It describes the ratio of the remaining frame's area after stabilization to the original one; 
{iv) \it Success Rate.} We also measure the success rate, which computes the ratio of videos that were successfully processed and yielded a distortion score lower than or equal to one.
% We introduce a new metric called success rate. While other metrics measure the quality of the video being stabilized, this metric measure the success rate.     \EN{Why are you creating a new metric? Give a brief motivation for that.}
% It presents the number of videos that a video stabilization algorithm successfully processed, to the total number of videos. A video with a distortion score higher than one is considered as a failure. 

\small{
\begin{table}[tb]
	\centering
	\caption{\textbf{Comparison in the Selfie Video dataset~\cite{yu2018selfie}}. Our method achieves the best distortion score and comparable results to other baselines. 	\textbf{Bold} indicates the best, \underline{underline} second best, and \textit{italic} the third best result. 
	}
% 	\EN{Consider including the citation each each method in the table.}
	\label{tab:results_selfie}
\begin{tabular}{lllll}
	\toprule %\hline
              Method               & \begin{tabular}[c]{@{}l@{}} Stability\\Avg. Score$\uparrow$\end{tabular} & \begin{tabular}[c]{@{}l@{}}Distortion\\ Score$\ast$\end{tabular} & \begin{tabular}[c]{@{}l@{}}Cropping\\ Ratio$\uparrow$\end{tabular} & \begin{tabular}[c]{@{}l@{}}Success\\ Rate$\uparrow$\end{tabular} \\
                             	\toprule %\hline
FuSta~\cite{liu2021hybrid}                & \underline{$0.818$}         & $\mathit{0.777}$    & $\mathbf{0.970}$   & $\mathbf{0.970}$ \\
Grundmann~\etal~\cite{grundmann2011auto}     & $0.727$         & \underline{$0.828$}    & $0.848$  & $0.848$ \\
StabNet~\cite{wang2018deep}              & $0.763$         & $0.680$    & $\mathit{0.917}$  & $0.667$ \\
DIFRINT~\cite{choi2020deep}              & $\mathbf{0.827}$        & $0.691$    & $0.912$  & $\mathit{0.915}$ \\
Yu~\etal~\cite{yu2020learning}             & $0.770$         & $0.739$    & $0.909$  & $0.909$ \\
Ours                 & $\mathit{0.787}$         & $\mathbf{0.933}$    & \underline{$0.939$}  & \underline{$0.939$} \\
			%\midrule
		    \multicolumn{5}{r}{\scriptsize{\textit{$\uparrow$Higher is better} \textit{$^\ast$Better closer to $1$}}} \\
			\bottomrule
\end{tabular}
\end{table}
}
\subsection{Results}
%In this section, we first show the quantitative results on VSAC105Real dataset. Then, we analyze the different design choices.  

%\paragraph*{Quantitative evaluation.}
% The stability average score of our method compared to five state-of-the-art video stabilization algorithms are shown in Table \ref{tab:results}.

Table~\ref{tab:results} shows the results of comparing our method to several video stabilization approaches. As can be seen, our method presented the best values on average in comparison to all the baselines in terms of stability average, distortion, cropping ratio, and success rate. 
% It is noteworthy that for the foggy weather condition, our method produced high-quality steady videos.
Even though our method did not surpass the baselines at each class individually, it still achieved competitive results. Every baseline performs badly in at least one class, while our method is more robust across classes, hence holding the final best results on the VSAC105Real. 

Preserving the content while compensating for camera shakiness is another important feature of our algorithm. Our method achieved the best results as compared to other state-of-the-art methods. The superiority of our method can be linked to the accurate affine transformation matrix estimation and the smoothing stage. Moreover, our method achieved the highest success rate compared to the competitors as shown in Table~\ref{tab:results}. It is worth noting that all baselines failed to stabilize most of the shaky videos at foggy weather conditions. The reason for this outcome is that participating media, like fog, work as a low pass filter that removes high-quality features that most videos stabilization algorithms depend on to estimate the camera trajectory. We highlight that even though our model was not trained on any samples under the foggy weather condition, it was capable of learning useful features from both raw images and optical flow.

Furthermore, we evaluate our model  on Selfie Video dataset~\cite{yu2018selfie}, contains videos under normal weather condition and standard illumination. Our model achieves comparable results to other methods while maintaining the best distortion score as shown in Table~\ref{tab:results_selfie}. It should be noted that our model was trained from scratch on synthetic data only.

\subsection{Ablation Study}
\setlength{\tabcolsep}{0.08em}
\small{
\begin{table}[t]
\centering
\caption{{\bf Ablation study.} Performance for different design choices (best in \textbf{bold}).}
\label{table:ablation}
% \resizebox{\columnwidth}{!}{%
\setlength{\tabcolsep}{1.0pt}
\begin{tabular}{lcccccc}
\toprule 
&
  \multicolumn{1}{c}{\begin{tabular}[c]{cc}Single\\Network\end{tabular}} &
  \multicolumn{1}{c}{SIFT} &
  \multicolumn{1}{c}{\begin{tabular}[c]{cc}More\\ Data\end{tabular}} &
  \multicolumn{1}{c}{\begin{tabular}[c]{cc}No Optical\\Flow\end{tabular}} &
  \multicolumn{1}{c}{\begin{tabular}[c]{cc}$l_{1}$ Directed\\Smoothing\end{tabular}} &
  \multicolumn{1}{c}{\begin{tabular}[c]{cc}Complete\\Model\end{tabular}} \\
  \midrule
Stability Avg. Score $\uparrow$       & $0.443$   & $0.576$   & $0.690$  & $0.678$ & $0.675$ & $ \mathbf{0.695}$ \tabularnewline  
Distortion Score $\ast$           & $0.540$   & $0.650$   & $0.793$  & $0.781$ & $0.828$ & $ \mathbf{0.845}$ \tabularnewline  
Cropping Ratio $\uparrow$             & $0.557$   & $0.670$   & $0.850$  & $0.829$ & $0.844$ & $ \mathbf{0.857}$  \tabularnewline  
Success Rate  $\uparrow$              & $0.543$   & $0.667$   & $0.829$  & $0.800$ & $ 0.838$ & $\mathbf{0.840}$  \tabularnewline  
%\midrule
\multicolumn{7}{r}{\scriptsize{\textit{$\uparrow$Higher is better} \textit{$^\ast$Better closer to $1$}}} \\
\bottomrule
\end{tabular}

% }
% \vspace{-3mm}
\end{table}
}

We analyzed different design options and showed the effectiveness of each component of our proposed pipeline. The results are reported in Table~\ref{table:ablation}. First, since we use two networks in our pipeline, we trained using a single CNN with four losses: horizontal translation loss, vertical translation loss, rotation loss, and scale loss. As a result, the network was unable to converge well. One problem could be that the translation losses were larger than others. However, even after applying weights for the losses to make them comparable to each other, the network could not learn well (column \textit{Single Network} in Table~\ref{table:ablation}).

To emphasize the advantages of using our learning-based model, for affine transformation matrix estimation over applying SIFT, we apply SIFT to find the affine transformation matrix while keeping the smoothing part of our model intact. As expected, SIFT did not perform well (column \textit{SIFT} in Table~\ref{table:ablation}). Standard feature extractors like SIFT struggle to extract reliable and robust features under adverse conditions. Rain and snow particles, low illumination at night, and foggy weather make finding and matching features rather hard. This leads to inaccurate affine transformations, thus low-quality stabilized videos. 
% To further evaluate the advantages of using our smoothing algorithm over $l_{1}$ directed smoothing as done in~\cite{grundmann2011auto}, we keep our learning-based model for affine transformation matrix estimation. However, we use $l_{1}$ directed smoothing on the predicted camera trajectory.
To further evaluate the advantages of using our smoothing algorithm over $l_{1}$ directed smoothing, as done in~\cite{grundmann2011auto}, we apply $l_{1}$ directed smoothing on the predicted camera trajectory, while keeping our learning-based model for the affine transformation matrix estimation. As expected, the model does not preform very well as compared to using our proposed smoothing algorithm (column \textit{$l_{1}$ directed smoothing} in Table~\ref{table:ablation}) because ours considers more sophisticated camera paths and is not limited to constant, linear and parabolic motions like~\cite{grundmann2011auto}.

To show the need for optical flow in the affine transformation learning, we train using only graysale images rather than both grayscale and optical flow modalities. As anticipated, not utilizing optical flow information decreased the video stabilization quality (column \textit{No Optical Flow} in Table~\ref{table:ablation}).

\begin{figure*}[t!]
    \centering
    \includegraphics[width=\textwidth]{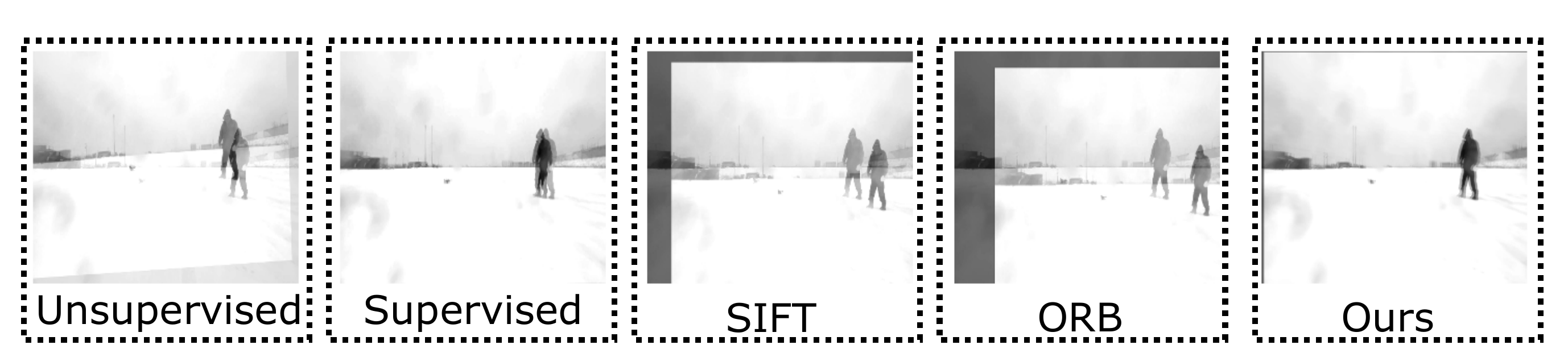}
    \caption{{\bf Qualitative comparison for affine transformation.} Comparison between unsupervised~\cite{zhang2020content}, supervised~\cite{detone2016deep}, SIFT~\cite{lowe2004distinctive}, ORB~\cite{rublee2011orb} and ours for affine transformation estimation. Our method can handle challenging conditions such as low textures where other methods perform poorly. }
    \label{fig:affine_est_qualit}
    % \vspace{-4mm}
\end{figure*}

At last, to demonstrate the effect of our training synthetic data on the quality of video stabilization, we trained our model on more data, including normal weather, adverse weathers, and night-time videos. The training was performed using VSAC65Synth dataset from scratch, and no real data were used. The results indicated no significant improvement on the baseline that was trained on VSNC35Synth dataset. Thus, we can see in Table~\ref{table:ablation}, under \textit{More Data} column, that a few synthetic videos, with accurate ground-truth, were sufficient to train our model to learn affine transformation. Thus, increasing the number and diversity of the training videos did not boost the overall performance.
\setlength{\tabcolsep}{0.15em}
\small{
\begin{table}[t]
\centering
	\caption{ \textbf{Affine matrix estimation}. Comparison among different methods for affine matrix estimation on CA-Unsupervised dataset~\cite{zhang2020content} using $l_{2}$~distance.
	\textbf{Bold} indicates the best result.
	} 
	\label{tab:affine_methods}
\begin{tabular}{lllcllcllcclcclcclc}
\toprule
Method &  &  & RE$\downarrow$   &  &  & LT$\downarrow$  &  &  & LL$\downarrow$   &   &  & SF$\downarrow$   &    &  & LF$\downarrow$  & &  & Average$\downarrow$  \\
\midrule
ORB~\cite{rublee2011orb} + RANSAC~\cite{fischler1981random}        &  &  & $9.24$               &  &  & $14.63$              &  &  & $12.27$              &                      &  & $11.36$              &                      &  & $7.20$               &                      &  & $10.94$              \\
ORB~\cite{rublee2011orb} + MAGSAC~\cite{barath2019magsac}          &  &  & $10.11$              &  &  & $19.79$              &  &  & $12.48$              &                      &  & $11.86$              &                      &  & $7.85$               &                      &  & $12.42$              \\
ORB~\cite{rublee2011orb} + LMEDS                                   &  &  & $9.78$               &  &  & $40.11$              &  &  & $12.02$              &                      &  & $10.84$              &                      &  & $7.01$               &                      &  & $15.95$              \\ 
\midrule
SIFT~\cite{lowe2004distinctive} + RANSAC~\cite{fischler1981random} &  &  & $10.63$              &  &  & $11.70$              &  &  & $13.37$              &                      &  & $11.75$              &                      &  & $6.44$               &                      &  & $10.78$              \\
SIFT~\cite{lowe2004distinctive} + MAGSAC~\cite{barath2019magsac}   &  &  & $10.75$              &  &  & $10.97$              &  &  & $12.99$              &                      &  & $11.09$              &                      &  & $6.35$               &                      &  & $10.43$              \\
SIFT~\cite{lowe2004distinctive} + LMEDS                            &  &  & $10.47$              &  &  & $9.72$               &  &  & $13.04$              &                      &  & $10.14$              &                      &  & $5.88$               &                      &  & $9.85$               \\ \midrule
Supervised~\cite{detone2016deep}                                   &  &  & $8.39$               &  &  & $9.33$               &  &  & $8.63$               &                      &  & $10.29$              &                      &  & $5.92$               &                      &  & $8.51$               \\
Unsupervised~\cite{zhang2020content}                               &  &  & $7.05$               &  &  & $7.60$               &  &  & $6.84$               &                      &  & $7.42$               &                      &  & $\textbf{3.84}$      &                      &  & $6.55$               \\ \midrule  
Ours                                                               &  &  & $\textbf{4.55}$      &  &  & $\textbf{5.58}$      &  &  & $\textbf{5.68}$      &                      &  & $\textbf{5.17}$      &                      &  & $9.73$               &                      &  & $\textbf{6.14}$      \\ 
% \midrule  
\multicolumn{1}{r}{}                                               &  &  & \multicolumn{1}{l}{} &  &  & \multicolumn{1}{l}{} &  &  & \multicolumn{1}{l}{} & \multicolumn{1}{l}{} &  & \multicolumn{1}{l}{} & \multicolumn{1}{l}{} &  & \multicolumn{1}{l}{} & \multicolumn{1}{l}{} &  & \multicolumn{1}{r}{\scriptsize{\textit{$\downarrow$Lower is better}}}\\
% \multicolumn{1}{r}{\scriptsize{\textit{$\downarrow$Lower is better}}}
% \textit{$\downarrow$Lower is better}
\bottomrule  

\end{tabular}
% \vspace{-6mm}
\end{table}
}
% \paragraph*{Our Affine Transformation Model.} 
To further investigate the accuracy of our estimated affine transformation matrix, we carry a special set of experiments shown in~Table~\ref{tab:affine_methods}. We compare our learning-based affine transformation estimation model to three types of affine estimation approaches: i) the traditional ones, including ORB and SIFT with RANSAC, MAGSAC, and LMEDS for outliers rejection; ii) supervised; and iii) unsupervised approaches. We note that traditional approaches can be used to estimate the affine transformation directly, but supervised and unsupervised methods are designed to estimate the homography matrix. Thus, for fair comparison, we extract the affine transformation from the estimated homography one. We utilize the dataset of~\cite{zhang2020content} which contains $4{,}200$ pairs of images where each image-pair includes 6-matching human-annotated pairs of points. The dataset covers regular-texture (RE), low-texture (LT), low-light (LL), small-foregrounds (SF), and large-foregrounds (LF). We use $l_{2}$ distance to measure the error between the warped and ground-truth points similar to~\cite{zhang2020content,ye2021motion}. Table~\ref{tab:affine_methods} demonstrates that our method can estimate the affine transformation better for both standard and challenging conditions. Figure~\ref{fig:affine_est_qualit} shows a qualitative comparison among our model, traditional (\eg, SIFT and ORB), supervised~\cite{detone2016deep} and unsupervised~\cite{zhang2020content} methods. While other methods fail under such challenging conditions, our method performs well because it learned how to extract resilient features using our specially designed synthetic data.

% \begin{table}[]
% \centering
% \caption{Ablations studies for different design choices.}
% \label{table:ablation}
% \resizebox{\columnwidth}{!}{%
% \begin{tabular}{lcccc}
% \hline
%  &
%   \multicolumn{1}{l}{\begin{tabular}[c]{@{}l@{}}Complete \\ Model\end{tabular}} &
%   \multicolumn{1}{l}{\begin{tabular}[c]{@{}l@{}}Single \\ Network\end{tabular}} &
%   \multicolumn{1}{l}{SIFT} &
%   \multicolumn{1}{l{\begin{tabular}[c]{@{}l@{}}More\\  Data\end{tabular}}} \\ \toprule
% Stability Score       & \textbf{0.730} & 0.507 & 0.593 & 0.714          \\ 
% Stability Rotation    & \textbf{0.738} & 0.482 & 0.596 & 0.698          \\ 
% Stability Translation & 0.721          & 0.531 & 0.591 & \textbf{0.731} \\ 
% Distortion Score      & \textbf{0.811} & 0.590 & 0.660 & 0.794          \\ 
% Cropping Ratio        & \textbf{0.864} & 0.595 & 0.684 & 0.858 \\ 
% Success Rate        & 0.836 & 0.586 & 0.680 & \textbf{0.844} \\ \bottomrule

% \end{tabular}
% }
% \end{table}
% \input{sections/limitations}
\section{Conclusion and Limitations}
We showed that most of the state-of-the-art video stabilization methods could not perform well under adverse weather conditions. We proposed a novel synthetic-aware video stabilization algorithm that requires only synthetic data for training. Our experimental results demonstrated that our method surpasses all other baselines. We also provided one real dataset (VSAC105Real) for video stabilization under adverse conditions and two synthetic datasets (VSNC35Synth and VSAC65Synth) for training purposes.
Our ablation studies demonstrated that current affine transformation matrix estimation methods fail under challenging conditions. Despite the fact that repetitive textures caused by rain and snowdrops, along with smoothing filters caused by fog and the absence of enough features at night altogether, pose certain challenges on current video stabilization algorithms, our video stabilization performed well under these challenges.

% Although our approach achieved state-of-the-art results, it has a few limitations. It can be trained only on synthetic data and does not require real-world videos. While this is an advantage, our algorithm, however, cannot leverage real data. In future work, we are planning to generalize our method to real data, too. 
% Our method achieves interesting results under foggy weather conditions. However, it only attains adequate performance in other classes. 
% \paragraph*{Limitations.}
Although our approach achieved state-of-the-art results, it has a few limitations. Increasing the number of synthetic training samples did not boost the performance of our video stabilization method. The reason could be that our model has already learned how to estimate the affine transformation matrix, and any further training videos were not advantageous. At the same time, our affine estimation model did not perform well under straightforward scenarios such as large-foreground images. Additionally, it achieved comparable results on selfie video dataset. That is due to the fact that our generated synthetic dataset did not include any selfie videos. Even though, our video stabilization algorithm obtained satisfactory results. Hence, we have demonstrated that synthetic data can be very useful, but not only as an additional data source, as how it was usually expected, but also as an essential element for novel synthetic-aware computer vision algorithms.    
% Thus, in the future we are planning to use active learning to guide the synthetic generation process. Thus, to generate a balanced dataset covering both standard and challenging cases. 

%Bibliography
\bibliographystyle{unsrt}  
\bibliography{main}

\begin{thebibliography}{10}

\bibitem{lowe2004distinctive}
David~G Lowe.
\newblock Distinctive image features from scale-invariant keypoints.
\newblock {\em International journal of computer vision}, 60(2):91--110, 2004.

\bibitem{r2d2}
Jerome Revaud, Philippe Weinzaepfel, C{\'{e}}sar~Roberto de~Souza, and Martin
  Humenberger.
\newblock {R2D2:} repeatable and reliable detector and descriptor.
\newblock In {\em NeurIPS}, 2019.

\bibitem{luo2020aslfeat}
Zixin Luo, Lei Zhou, Xuyang Bai, Hongkai Chen, Jiahui Zhang, Yao Yao, Shiwei
  Li, Tian Fang, and Long Quan.
\newblock Aslfeat: Learning local features of accurate shape and localization.
\newblock {\em Computer Vision and Pattern Recognition (CVPR)}, 2020.

\bibitem{liu2021urbanscene3d}
Yilin Liu, Fuyou Xue, and Hui Huang.
\newblock Urbanscene3d: A large scale urban scene dataset and simulator.
\newblock {\em arXiv preprint arXiv:2107.04286}, 2021.

\bibitem{kerim2021using}
Abdulrahman Kerim, Ufuk Celikcan, Erkut Erdem, and Aykut Erdem.
\newblock Using synthetic data for person tracking under adverse weather
  conditions.
\newblock {\em Image and Vision Computing}, 111:104187, 2021.

\bibitem{shafaei2016play}
Alireza Shafaei, James~J Little, and Mark Schmidt.
\newblock {Play and Learn: Using Video Games to Train Computer Vision Models}.
\newblock {\em arXiv:1608.01745}, 2016.

\bibitem{tsirikoglou2022synthetic}
Apostolia Tsirikoglou.
\newblock {\em Synthetic data for visual machine learning: A data-centric
  approach}.
\newblock PhD thesis, Link{\"o}ping University Electronic Press, 2022.

\bibitem{wang2018deep}
Miao Wang, Guo-Ye Yang, Jin-Kun Lin, Song-Hai Zhang, Ariel Shamir, Shao-Ping
  Lu, and Shi-Min Hu.
\newblock Deep online video stabilization with multi-grid warping
  transformation learning.
\newblock {\em IEEE Transactions on Image Processing}, 28(5):2283--2292, 2018.

\bibitem{liu2021hybrid}
Yu-Lun Liu, Wei-Sheng Lai, Ming-Hsuan Yang, Yung-Yu Chuang, and Jia-Bin Huang.
\newblock Hybrid neural fusion for full-frame video stabilization.
\newblock {\em arXiv preprint arXiv:2102.06205}, 2021.

\bibitem{yu2020learning}
Jiyang Yu and Ravi Ramamoorthi.
\newblock Learning video stabilization using optical flow.
\newblock In {\em Proceedings of the IEEE/CVF Conference on Computer Vision and
  Pattern Recognition}, pages 8159--8167, 2020.

\bibitem{ali2020learning}
Muhammad~Kashif Ali, Sangjoon Yu, and Tae~Hyun Kim.
\newblock Learning deep video stabilization without optical flow.
\newblock {\em arXiv preprint arXiv:2011.09697}, 2020.

\bibitem{kerim2021silver}
Abdulrahman Kerim, Leandro Soriano~Marcolino, and Richard Jiang.
\newblock Silver: Novel rendering engine for data hungry computer vision
  models.
\newblock In {\em 2nd International Workshop on Data Quality Assessment for
  Machine Learning}, 2021.

\bibitem{grundmann2011auto}
Matthias Grundmann, Vivek Kwatra, and Irfan Essa.
\newblock Auto-directed video stabilization with robust l1 optimal camera
  paths.
\newblock In {\em CVPR 2011}, pages 225--232. IEEE, 2011.

\bibitem{bradley2021cinematic}
Arwen Bradley, Jason Klivington, Joseph Triscari, and Rudolph van~der Merwe.
\newblock {Cinematic-L1 Video Stabilization with a Log-Homography Model}.
\newblock In {\em Proceedings of the IEEE/CVF Winter Conference on Applications
  of Computer Vision}, pages 1041--1049, 2021.

\bibitem{choi2020deep}
Jinsoo Choi and In~So Kweon.
\newblock Deep iterative frame interpolation for full-frame video
  stabilization.
\newblock {\em ACM Transactions on Graphics (TOG)}, 39(1):1--9, 2020.

\bibitem{rublee2011orb}
Ethan Rublee, Vincent Rabaud, Kurt Konolige, and Gary Bradski.
\newblock {ORB: An efficient alternative to SIFT or SURF}.
\newblock In {\em 2011 International conference on computer vision}, pages
  2564--2571. Ieee, 2011.

\bibitem{bay2006surf}
Herbert Bay, Tinne Tuytelaars, and Luc Van~Gool.
\newblock {SURF: Speeded up robust features}.
\newblock In {\em European conference on computer vision}, pages 404--417.
  Springer, 2006.

\bibitem{zhang2019learning}
Jiahui Zhang, Dawei Sun, Zixin Luo, Anbang Yao, Lei Zhou, Tianwei Shen, Yurong
  Chen, Long Quan, and Hongen Liao.
\newblock Learning two-view correspondences and geometry using order-aware
  network.
\newblock In {\em Proceedings of the IEEE/CVF International Conference on
  Computer Vision}, pages 5845--5854, 2019.

\bibitem{fischler1981random}
Martin~A Fischler and Robert~C Bolles.
\newblock Random sample consensus: a paradigm for model fitting with
  applications to image analysis and automated cartography.
\newblock {\em Communications of the ACM}, 24(6):381--395, 1981.

\bibitem{barath2019magsac}
Daniel Barath, Jiri Matas, and Jana Noskova.
\newblock {MAGSAC: marginalizing sample consensus}.
\newblock In {\em Proceedings of the IEEE/CVF Conference on Computer Vision and
  Pattern Recognition}, pages 10197--10205, 2019.

\bibitem{zhang2020content}
Jirong Zhang, Chuan Wang, Shuaicheng Liu, Lanpeng Jia, Nianjin Ye, Jue Wang,
  Ji~Zhou, and Jian Sun.
\newblock Content-aware unsupervised deep homography estimation.
\newblock In {\em European Conference on Computer Vision}, pages 653--669.
  Springer, 2020.

\bibitem{detone2016deep}
Daniel DeTone, Tomasz Malisiewicz, and Andrew Rabinovich.
\newblock Deep image homography estimation.
\newblock {\em arXiv preprint arXiv:1606.03798}, 2016.

\bibitem{ye2021motion}
Nianjin Ye, Chuan Wang, Haoqiang Fan, and Shuaicheng Liu.
\newblock Motion basis learning for unsupervised deep homography estimation
  with subspace projection.
\newblock In {\em Proceedings of the IEEE/CVF International Conference on
  Computer Vision}, pages 13117--13125, 2021.

\bibitem{li2015dual}
Shiwei Li, Lu~Yuan, Jian Sun, and Long Quan.
\newblock Dual-feature warping-based motion model estimation.
\newblock In {\em Proceedings of the IEEE International Conference on Computer
  Vision}, pages 4283--4291, 2015.

\bibitem{butler2012naturalistic}
Daniel~J Butler, Jonas Wulff, Garrett~B Stanley, and Michael~J Black.
\newblock {A Naturalistic Open Source Movie for Optical Flow Evaluation}.
\newblock In {\em European conference on computer vision}, 2012.

\bibitem{richter2016playing}
Stephan~R Richter, Vibhav Vineet, Stefan Roth, and Vladlen Koltun.
\newblock {Playing for Data: Ground Truth from Computer Games}.
\newblock In {\em European conference on computer vision}, 2016.

\bibitem{dosovitskiy2017carla}
Alexey Dosovitskiy, German Ros, Felipe Codevilla, Antonio Lopez, and Vladlen
  Koltun.
\newblock {CARLA: An open urban driving simulator}.
\newblock In {\em Conference on robot learning}, pages 1--16. PMLR, 2017.

\bibitem{hao2021gancraft}
Zekun Hao, Arun Mallya, Serge Belongie, and Ming-Yu Liu.
\newblock {GANcraft: Unsupervised 3D Neural Rendering of Minecraft Worlds}.
\newblock {\em arXiv preprint arXiv:2104.07659}, 2021.

\bibitem{lee2009video}
Ken-Yi Lee, Yung-Yu Chuang, Bing-Yu Chen, and Ming Ouhyoung.
\newblock Video stabilization using robust feature trajectories.
\newblock In {\em 2009 IEEE 12th International Conference on Computer Vision},
  pages 1397--1404. IEEE, 2009.

\bibitem{liu2012video}
Shuaicheng Liu, Yinting Wang, Lu~Yuan, Jiajun Bu, Ping Tan, and Jian Sun.
\newblock Video stabilization with a depth camera.
\newblock In {\em 2012 IEEE Conference on Computer Vision and Pattern
  Recognition}, pages 89--95. IEEE, 2012.

\bibitem{IMKDB17}
E.~Ilg, N.~Mayer, T.~Saikia, M.~Keuper, A.~Dosovitskiy, and T.~Brox.
\newblock Flownet 2.0: Evolution of optical flow estimation with deep networks.
\newblock In {\em IEEE Conference on Computer Vision and Pattern Recognition
  (CVPR)}, Jul 2017.

\bibitem{flownet2-pytorch}
Fitsum Reda, Robert Pottorff, Jon Barker, and Bryan Catanzaro.
\newblock {Flownet2-pytorch: Pytorch implementation of FlowNet 2.0: Evolution
  of Optical Flow Estimation with Deep Networks}.
\newblock \url{https://github.com/NVIDIA/flownet2-pytorch}, 2017.

\bibitem{savitzky1964smoothing}
Abraham Savitzky and Marcel~JE Golay.
\newblock Smoothing and differentiation of data by simplified least squares
  procedures.
\newblock {\em Analytical chemistry}, 36(8):1627--1639, 1964.

\bibitem{gonzalez2020rocketbox}
Mar Gonzalez-Franco, Ofek, et~al.
\newblock {The Rocketbox Library and the Utility of Freely Available Rigged
  Avatars}.
\newblock {\em Frontiers in virtual reality}, 1(article 561558), 2020.

\bibitem{zhang2018full}
Lei Zhang, Qing-Zhuo Zheng, Hong-Kang Liu, and Hua Huang.
\newblock Full-reference stability assessment of digital video stabilization
  based on riemannian metric.
\newblock {\em IEEE Transactions on Image Processing}, 27(12):6051--6063, 2018.

\bibitem{yu2018selfie}
Jiyang Yu and Ravi Ramamoorthi.
\newblock Selfie video stabilization.
\newblock In {\em Proceedings of the European Conference on Computer Vision
  (ECCV)}, pages 551--566, 2018.

\bibitem{liu2013bundled}
Shuaicheng Liu, Lu~Yuan, Ping Tan, and Jian Sun.
\newblock Bundled camera paths for video stabilization.
\newblock {\em ACM Transactions on Graphics (TOG)}, 32(4):1--10, 2013.

\bibitem{kingma2014adam}
Diederik~P Kingma and Jimmy Ba.
\newblock Adam: A method for stochastic optimization.
\newblock {\em arXiv preprint arXiv:1412.6980}, 2014.

\end{thebibliography}

\end{document}